\documentclass[letterpaper, journal]{IEEEtran}

\usepackage{mathtools}    
\usepackage{amssymb}
\usepackage{amsfonts}

\usepackage{graphicx}
\usepackage[dvipsnames,table]{xcolor}

\usepackage{cite}
\usepackage{url}

\usepackage{booktabs}
\usepackage{array}
\usepackage{tabularx}
\usepackage{placeins}
\usepackage{makecell}
\usepackage{siunitx}
\newcolumntype{Y}{>{\centering\arraybackslash}X}

\usepackage[caption=false,font=footnotesize]{subfig}

\usepackage{algorithm}
\usepackage{algorithmic}

\usepackage{textcomp}
\usepackage{verbatim}
\usepackage[utf8]{inputenc}
\usepackage{balance}
\usepackage{float}
\usepackage{listings}
\usepackage[normalem]{ulem}
\usepackage{pifont}
\usepackage{tcolorbox}
\tcbuselibrary{listings,breakable,skins}
\usepackage{lipsum}
\usepackage{longtable}
\usepackage{seqsplit}
\newif\ifshowcomment
\newif\ifnumberrevision
\newif\ifcolorrevision
\newif\ifstrikeremovel

\showcommenttrue
\numberrevisiontrue
\colorrevisiontrue
\strikeremoveltrue

\usepackage{tcolorbox}
\tcbuselibrary{breakable}
\usepackage{ragged2e}
\usepackage{setspace}

\newtcolorbox{promptbox}{
  enhanced,
  breakable,
  colback=gray!3,
  colframe=black!20,
  boxrule=0.35pt,
  arc=0.8mm,
  outer arc=0.8mm,
  left=3pt,
  right=3pt,
  top=3pt,
  bottom=3pt,
  before upper={\scriptsize\setstretch{0.84}\justifying},
}

\newcommand{\jsonline}[1]{%
  {\ttfamily\scriptsize\RaggedRight\detokenize{#1}\par}%
}

\usepackage{xcolor}

\newcommand{\rev}[1]{\textcolor{black}{#1}}
\newcommand{\revtwo}[1]{\textcolor{black}{#1}}

\newenvironment{NoteToPractitioners}
{%
  \par\vspace{0.8ex}%
  \noindent\small%
  \textbf{\textit{Note to Practitioners---}}%
  \bfseries\ignorespaces%
}
{%
  \par\vspace{0.8ex}\normalfont\normalsize%
}

\title{From Perception to Symbolic Task Planning: Vision-Language Guided Human-Robot Collaborative Structured Assembly}

\author{Yanyi \author{Yanyi Chen, Min Deng$^{*}$

\thanks{$^{*}$Corresponding author: Min Deng.}

\thanks{Yanyi Chen and Min Deng were with the Department of Civil, Environmental, and Construction Engineering, Texas Tech University, Lubbock, TX 79409, USA. They are now with the Department of Civil and Environmental Engineering, University of Tennessee, Knoxville, TN 37996, USA (e-mail: yanychen@utk.edu; mindeng@utk.edu).}
}
}

\begin{document}
\maketitle

\begin{abstract}
\rev{Despite recent advances in automation, many real-world assembly tasks still require close collaboration between humans and robots due to the complexity, uncertainty, and safety constraints of dynamic and unstructured environments.}  Human-robot collaboration (HRC) in structured assembly requires reliable state estimation and adaptive task planning under noisy perception and human interventions. To address these challenges, we introduce a design-grounded human-aware planning framework for human-robot collaborative structured assembly. The framework comprises two coupled modules. Module I, Perception-to-Symbolic State (PSS), employs vision-language models (VLMs) based agents to align RGB-D observations with design specifications and domain knowledge, synthesizing verifiable symbolic assembly states. It outputs validated installed and uninstalled component sets for online state tracking. Module II, Human-Aware Planning and Replanning (HPR), performs task-level multi-robot assignment and updates the plan only when the observed state deviates from the expected execution outcome. It applies a minimal-change replanning rule to selectively revise task assignments and preserve plan stability even under human interventions. 
We validate the framework on a 27-component timber-frame assembly. The PSS module achieves \textbf{97\%} state synthesis accuracy, and the HPR module maintains feasible task progression across diverse HRC scenarios.
Results indicate that integrating VLM-based perception with knowledge-driven planning improves robustness of state estimation and task planning under dynamic conditions.
\end{abstract}

\begin{NoteToPractitioners}
Robotic systems operating in structured assembly environments must adapt to frequent state changes caused by perception uncertainty and unexpected human intervention. Practitioners often struggle to maintain an accurate representation of assembly progress while updating task plans without disrupting ongoing work. This work presents a design-grounded planning framework that supports human-robot collaborative assembly at the task level. The approach leverages existing design artifacts and perception outputs to maintain a consistent symbolic representation of installed and uninstalled components. When deviations occur due to human intervention, task assignments are selectively updated using a minimal-change strategy to preserve plan stability and workflow continuity. The framework does not require modifications to low-level robot control and can be integrated with existing multi-robot and human-robot collaborative construction systems. While validated in a timber-frame assembly scenario, the method is applicable to other structured assembly tasks involving HRC.
\end{NoteToPractitioners}


\section{Introduction}

\IEEEPARstart{S}{tructured} assembly, such as the construction of timber-frame building components, presents a significant challenge for robotic automation~\cite{bockFutureConstructionAutomation2015}. While robots excel at repetitive tasks in controlled environments, their deployment in the real world remains challenging due to dynamic site conditions, execution failures, and unscripted human interventions~\cite{sicilianoRobotics2009}. Unlike fully scripted industrial tasks, structured assembly tasks are characterized by evolving partial structures, strong precedence constraints, and incomplete observability during execution~\cite{ranjandasSustainableManufacturingReview2025}. Components are installed incrementally, and the physical configuration of the structure changes continuously as assembly progresses. Moreover, humans work alongside robots and may perform unrecorded actions, such as temporarily fixing or installing components, further increasing uncertainty in the system state~\cite{ajoudaniProgressProspectsHuman2018}.

The first challenge concerns robust state estimation, which requires inferring the assembly state from real-world observations.
In structured assembly, the state does not refer to precise geometric poses, but to a component-level understanding of which parts have been installed, which remain uninstalled, and how the current configuration relates to the design specification.
However, visual observations are often noisy and incomplete due to occlusions by workers, temporary fixtures, partial views of large structures, and sensing artifacts, making it difficult to reliably infer discrete assembly states from raw perception~\cite{tenorthKnowRobKnowledgeProcessing2013,homemdemelloRepresentationsMechanicalAssembly1991}.
The second challenge concerns adaptive task planning under structured assembly constraints.
Based on the current symbolic state, the system must dynamically generate and revise actions for a heterogeneous human-robot team while preserving the structural validity of the assembly.
In structured assembly, task admissibility and ordering are governed by domain-specific construction rules, such as support dependencies and precedence relations defined in the design specification~\cite{gerkeyFormalAnalysisTaxonomy2004,ajoudaniProgressProspectsHuman2018}. Human interventions can violate these assumptions by completing components out of order, removing supporting elements, or modifying the set of admissible components, thereby invalidating previously feasible plans.
Naïve replanning strategies that regenerate plans from scratch ignore these structural dependencies and may unnecessarily reassign unaffected tasks, disrupt ongoing collaboration, and reduce plan stability in human-robot teams.
These challenges lead to the following research questions:
\begin{itemize}
    \item \textbf{RQ1:} How can a robotic system reconcile ambiguous and partially visual observations with a predefined design specification to produce a verifiable, component-level symbolic assembly state that is consistent with structural and construction constraints?
    \item \textbf{RQ2:} How can a task-level planning framework update and preserve coordinated human-robot assembly actions under unscripted human interventions, while respecting design-defined structural dependencies and minimizing unnecessary disruption to unaffected tasks?
\end{itemize}

To address these questions, we propose a closed-loop planning framework that combines design-grounded symbolic state synthesis with human-aware task planning. Human awareness in this paper is defined at the task and symbolic state level, where human actions are treated as exogenous events that modify the assembly state and trigger selective replanning.
The framework maintains consistency between perceptual observations, design specifications, and planning decisions under perceptual uncertainty and unscripted human interventions.


The main contributions of this work are summarized as follows:

\begin{enumerate}
    \item We present a closed-loop, design-grounded planning framework that formalizes the interaction between symbolic state synthesis and human-aware task planning for robust human-robot collaborative assembly.

    \item We propose a design-grounded symbolic state synthesis method that converts static design artifacts (e.g., DXF) into structured symbolic representations. By aligning vision–language model (VLM) inference with explicit engineering constraints, the method enables automated component-level state reasoning without manual semantic labeling.

    \item We introduce a knowledge-driven human-aware task planning approach that determines valid next assembly actions and selectively updates task assignments in response to unscripted human interventions, without requiring full replanning.

    \item We validate the proposed framework through a 27-component timber-frame wall assembly case study, demonstrating high symbolic state accuracy and reliable task progression under dynamic human-robot collaboration (HRC) scenarios.
\end{enumerate}


The remainder of this paper is organized as follows. 
Section~\ref{sec:RW} reviews related work on knowledge-based assembly, foundation models in robotics, and human-robot collaborative planning. 
Section~\ref{sec:method} presents the developed methodology, followed by the case study and experimental results in
Section~\ref{sec:case}. 
Section~\ref{sec:discussion} discusses implications and limitations, and Section~\ref{sec:con} concludes the paper.

\section{Related Work}
\label{sec:RW}

\subsection{Assembly Planning from Design Specifications}

Assembly planning from design specifications extracts task representations from predefined models and knowledge structures, such as CAD files, graphical instructions, ontologies, and behavioral taxonomies.
CAD-based approaches derive task representations from geometric and relational information in three-dimensional models, for example, by extracting part dependencies through liaison graph construction~\cite{zobov2023autoassemblyframeworkautomatedrobotic}.
While they provide precise geometric information, they require complete CAD models and cannot handle assemblies documented only through visual instructions or manuals.

To overcome the reliance on complete CAD models, Wang et al.~\cite{errorcorrection2023} propose extracting assembly task sequence graphs directly from graphical instruction manuals, enabling task reasoning when explicit design models are unavailable.
Building on this direction, Lv et al.~\cite{scenegraph2024} extend manual-based parsing to multimodal scene graph representations that encode spatial relations among agents, objects, and assembly components.

Beyond visual parsing, explicit knowledge representations have been used to formalize assembly semantics and support symbolic planning.
Zhao et al.~\cite{zhao2024ontology} model products, processes, and resources using an intermediate engineering ontology that is translated into Planning Domain Definition Language (PDDL) for multi-agent task scheduling.
In contrast, Angleraud et al.~\cite{knowledgebased2025} adopt ontology-grounded Hierarchical Task Networks (HTN) to generate collaborative assembly plans under predefined operators and constraints.

At a higher level, taxonomy and knowledge-based abstractions have been proposed to structure human-robot assembly skills. Lee et al. \cite{LEE2024102686} introduce the human-robot shared assembly taxonomy framework. It decomposes tasks into primitive tasks and atomic actions to facilitate knowledge transfer between humans and robots. Conti et al. \cite{commonsense2022} incorporate commonsense knowledge to optimize task allocation based on safety and ergonomic constraints. To support skill generalization, Sun et al. \cite{skillrepresentation2025} propose a multi-layer, multi-level knowledge representation structured across task, area, object, action, and agent layers. They use a knowledge-enhanced task-to-action model for manipulation reasoning and skill transfer, while learning-based approaches reduce manual modeling by deriving executable task representations from demonstrations, including instructional videos~\cite{learningbased2020}.

Despite differences in representation formalism, these methods share a common assumption that task representations derived from design artifacts align with physical assembly states. 
When physical configurations deviate from specifications, these frameworks cannot resolve the mismatch, often requiring manual intervention or replanning from scratch.

\subsection{Foundation Models in Assembly}

\rev{To address the reliance on complete design models and manual semantic labeling, foundation models have been introduced into robotic assembly to improve flexibility in task planning, typically by augmenting specific stages of the pipeline rather than replacing the full planning process.}

Several studies employ Large Language Models (LLMs) for high-level task structure generation. For example, 
Bashir et al.~\cite{wang2023stateexplore} use LLMs to map visual observations to symbolic task states and predict subsequent assembly steps, while Macaluso et al.~\cite{macaluso2024chatgptassembly} leverage CAD-derived metadata to guide LLM-based generation of robot control programs. 
Ao et al.~\cite{ao2025llmasbt} further structure task execution by using behavior trees as an explicit control representation.

Other studies ground symbolic reasoning in physical environments to better handle real-world complexity.
Specifically, Rana et al.~\cite{rana2023sayplan} combine LLM-based semantic reasoning with hierarchical 3D scene graphs, enabling task planning that explicitly accounts for spatial structure.
Similarly, You et al.~\cite{you2023robogpt} focus on assembly scenarios and infer task sequences under explicit spatial and resource constraints using symbolic planning.
Forlini et al.~\cite{drmgpt2024} integrate multimodal perception to identify component states and update task plans when human interventions modify the workspace.

Beyond task generation, recent work addresses adaptation during execution to handle uncertainty. 
Kang et al.~\cite{huang2023contingency}, for instance, extend HTN planning by translating natural language instructions into executable procedures, enabling contingency handling at the task level.
Moreover, Hao et al.~\cite{hao2025learngenplan} propose the Learn-Gen-Plan framework, which leverages learned motion primitives and perception feedback to support long-horizon task adaptation.
In construction contexts, Deng et al.~\cite{deng2025integratingllmsdigitaltwins} integrate LLMs with Digital Twins to translate unstructured narrative site updates into constraint changes for multi-robot task allocation (MRTA).

In general, foundation models improve task adaptability in assembly planning. 
However, most existing methods assume that the perceived state used for planning is sufficiently accurate, without explicitly verifying consistency between perception outputs and the physical assembly state.
Under incomplete or occluded observations, this assumption can lead to undetected mismatches between inferred task progress and actual execution, with limited mechanisms for correction.

\subsection{Human-Robot Collaborative Planning}

While foundation models improve task representation adaptability, most planning frameworks assume a controllable workspace. Human collaborators introduce additional uncertainty through unscripted actions, motivating behavioral modeling and real-time adaptation. Early work has focused on modeling human behavior to enable robots to anticipate and complement human interventions. 
Representative approaches include user-aware hierarchical task planning to monitor human progress and select robot actions without explicit communication~\cite{ramachandruni2023uhtp,pegoraro2025euhtp}.
To manage the unpredictable nature of human behavior, Cramer et al. \cite{xu2024probabilistic} applied Partially Observable Markov Decision Processes to estimate assembly intentions from state graphs. 
Similarly, Nemlekar et al. \cite{transferpref2023} used transfer learning to adjust robot assistance based on individual human habits observed in earlier tasks. 
While these methods enable intention inference and responsive robot behavior, they primarily focus on behavioral prediction and do not explicitly address task distribution among multiple agents.

Other studies focus on task allocation between humans and robots to balance efficiency with physical and geometric constraints. 
Examples include task allocation methods that consider part stability and geometric complexity~\cite{assemblyallocation2024},
PDDL-based assignment considering reachability and safety~\cite{chen2024optimaltaskplanningagentaware}, and discrete-event coordination using event-driven timing models~\cite{giacomuzzo2024decafdiscreteeventbasedcollaborative}. 
Although these approaches improve coordination, task assignments are typically computed based on an initial workspace configuration and provide limited mechanisms for revision when the assembly state changes during execution.

To improve robustness during execution, several frameworks introduce reactive and adaptive planning. El Makrini et al. \cite{hfsm2022} used a hierarchical finite-state machine to reassign tasks based on the operator's current workload and physical strain. 
For real-time adaptation, Zhou et al. \cite{temporal2023} developed a linear temporal logic planner that allows robots to respond to local changes using eye-in-hand cameras. 
Pupa et al. \cite{modularhrc2022} proposed a modular architecture that separates high-level scheduling from low-level trajectory planning to improve safety.

Nevertheless, most existing methods do not verify whether perceived assembly states satisfy the assembly constraints.
Without such verification, unexpected human actions or sensing noise can lead planners to operate on outdated or incorrect states, causing errors that propagate into execution and often require manual intervention to resolve.

\section{Methodology}
\label{sec:method}

To address the existing gaps, we propose a design-grounded human-aware planning framework for multi-robot structured assembly. We formalize the planning problem by defining symbolic assembly states, admissible component frontiers, and a minimal-change rule for plan updates under human intervention. The resulting architecture consists of a symbolic state synthesis module and a human-aware planning module. Table~\ref{tab:symbols} summarizes the notation used in the following sections.

\begin{table}[!t]
\caption{Definition of notation.}
\label{tab:symbols}
\centering
\small
\renewcommand{\arraystretch}{1.25}
\setlength{\tabcolsep}{4.5pt}
\begin{tabular}{p{0.12\columnwidth} p{0.8\columnwidth}}
\hline
\textbf{Symbol} & \textbf{Description} \\
\hline
$\mathcal{D}$ & Design artifact (e.g., CAD, BIM, or PDF) \\
$V$ & Set of design components \\
$\mathbf{A}$ & Multidimensional component array extracted from $\mathcal{D}$ \\
$I^{\mathrm{des}}$ & Design image rendered from $\mathcal{D}$ \\
$O_t$ & RGB-D observation at iteration $t$, consisting of $I_t^{RGB}$ and $I_t^{Depth}$ \\
$I_t^{RGB}$ & RGB images at iteration $t$ \\
$I_t^{Depth}$ & Depth images at iteration $t$ \\
$\tilde{I}_t^{\mathrm{perc}}$ & Design-aligned perceptual representation \\
$U_t^{raw}$ & Symbolic hypothesis of the uninstalled component set \\
$I_t$ & Installed component set at iteration $t$ \\
$U_t$ & Uninstalled component set at iteration $t$ \\
$\Gamma_{\mathrm{rules}}$ & Rule-based reconciliation operator \\
$\mathcal{R}$ & Heterogeneous robot team \\
$\mathcal{H}$ & Set of human collaborators \\
$C_j$ & Capability profile of robot $r_j$ \\
$T_t$ & Frontier of admissible components at iteration $t$ \\
$\Gamma_i^t$ & Robot assignment for component $v_i$ at iteration $t$ \\
$P_t$ & Task assignment plan at iteration $t$ \\
$\mathcal{E}_t$ & Event flag indicating a detected state inconsistency \\
$\mathcal{B}_t^{k}$ & \rev{Behavior Tree (BT)} executed by robot $r_k$ at iteration $t$ \\

\hline
\end{tabular}
\end{table}



\begin{figure*}[!t]
    \centering
    \includegraphics[width=\textwidth]{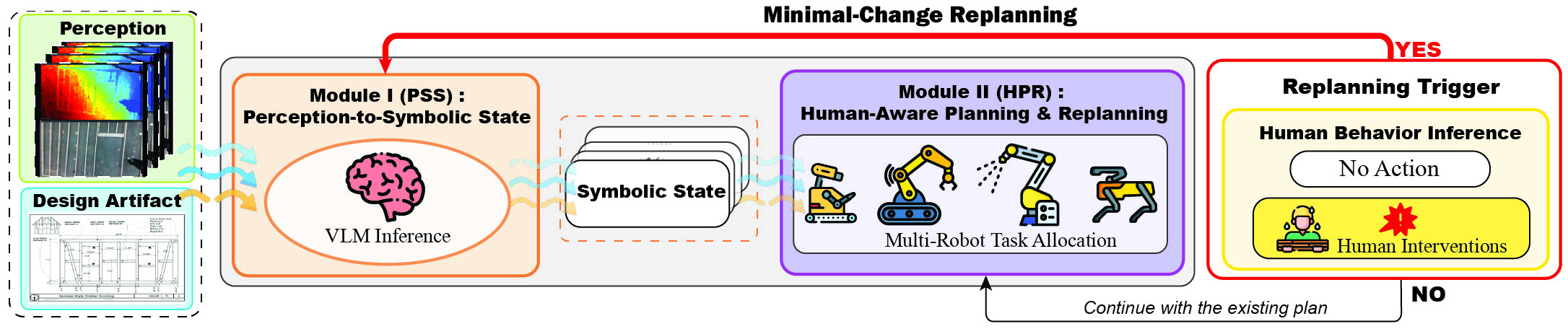}
    \caption{\rev{Overview of the proposed planning framework, showing the
    interaction between the Perception-to-Symbolic State module and
    Human-Aware Planning and Replanning module in a closed-loop
    perception-to-task-planning process.}}
    \label{fig:pipeline}
\end{figure*}

\subsection{Problem Formulation}
\label{sec:problem_formulation}

The structured assembly task is specified by a design artifact $\mathcal{D}$, which defines the components, their geometric and topological relations, and the required installation order.
In practice, the artifact is commonly available in formats such as CAD, BIM, or PDF.
Although the file format may differ across domains, the encoded design information is equivalent.
During execution, the workspace state changes over time due to robot actions, human interventions, and environmental variation, requiring continuous state updates and replanning.

\subsubsection{\textbf{Entities and inputs}}

Let $V = \{v_1, \dots, v_N\}$ denote the set of structural components specified by a design artifact $\mathcal{D}$. 
The robotic team is denoted by $\mathcal{R} = \{r_1, \dots, r_K\}$. 
Each robot $r_j$ is associated with a capability profile $C_j$, which encodes payload limits, kinematic reach, end-effector type, and workspace constraints. 
Human collaborators are represented by a set $\mathcal{H}$ of uncontrollable but observable agents.

At iteration $t$, the system obtains an observation $O_t$ and classifies each component in $V$ as installed or uninstalled, forming $I_t$ and $U_t$. 
Changes in these sets reflect both robot execution and possible human intervention. 
Human interventions are not modeled explicitly and are incorporated only through updates to the perceived state.

\subsubsection{\textbf{Problem statement}}

Each component $v_i \in V$ is associated with a prerequisite set $Pre(v_i)$ derived from the design artifact $\mathcal{D}$. 
The set $Pre(v_i)$ encodes installation precedence relations specified by the design and is assumed to be known.

At planning iteration $t$, the admissible frontier $T_t$ consists of all uninstalled components whose prerequisites are fully installed according to the verified symbolic state:
\begin{equation}
T_t = \{\, v_i \in U_t \mid Pre(v_i)\subseteq I_t \,\}.
\label{eq:frontier}
\end{equation}
A plan at iteration $t$ assigns each component in $T_t$ to a feasible robot coalition responsible for its installation. An assignment is feasible only if the selected robots satisfy all capability constraints for the component. 
These constraints include payload limits, kinematic reach, collision safety, and awareness of perceived human occupancy.
All constraints are evaluated using the current observation $O_t$.

Human-triggered events are detected when changes in the uninstalled set cannot be explained by robot execution. 
Such discrepancies are recorded as an event flag $\mathcal{E}_t$, indicating that the current plan may no longer be consistent with the updated state. If no human-triggered event is detected, the previous plan is retained. 
Otherwise, the planner selects a new feasible plan that minimizes the plan-edit distance $d(P_t,P_{t-1})$:
\begin{equation}
P_t =
\begin{cases}
P_{t-1}, & \mathcal{E}_t=\textsc{None}, \\[4pt]
\displaystyle \arg\min_{P_t} d(P_t,P_{t-1}), & \mathcal{E}_t\neq\textsc{None}.
\end{cases}
\label{eq:minchange}
\end{equation}

A plan $P_t$ specifies, for each admissible component $v_i \in T_t$, a set of robots assigned to install it, denoted by $\Gamma_i^t \subseteq \mathcal{R}$. 
The plan-edit distance is defined as the number of components whose assigned robots change between iterations:
\begin{equation}
d(P_t,P_{t-1})
=
\big|\{\, v_i \in T_t\cap T_{t-1} \mid \Gamma_i^t \ne \Gamma_i^{t-1} \}\big|.
\label{eq:edit_distance}
\end{equation}

\subsection{Framework architecture}
\label{sec:architecture}

Building on the problem formulation in Sec.~\ref{sec:problem_formulation}, the proposed framework defines a closed-loop architecture for HRC structured assembly. As illustrated in Fig.~\ref{fig:pipeline}, the architecture integrates the static design specification, live RGB-D observations, and domain knowledge into a unified semantic planning framework that supports adaptive multi-agent coordination.

\textbf{Module I, Perception-to-Symbolic State (PSS)} converts raw perceptual data $O_t=\{I_t^{RGB},I_t^{Depth}\}$ and the design artifact $\mathcal{D}$ into verified symbolic states of the current assembly procedure. It aligns the design geometry with the physical scene, performs VLM-based semantic matching, and applies deterministic rule-based reconciliation to produce the uninstalled component set $U_t$ for downstream planning. 

\textbf{Module II, Human-Aware Planning and Replanning (HPR)} takes these symbolic states as input and generates the corresponding multi-robot assembly plan $P_t$. It identifies the admissible frontier of installable components, determines feasible robot coalitions based on physical capabilities, and compiles the resulting assignments into per-robot behavior trees $\mathcal{B}_t^{k}$ that support parallel execution and synchronized collaborative actions.

After each iteration, the PSS module updates the assembly state. 
If the observed progression matches the expected evolution, the system continues executing the current plan. When human interventions lead to changes in the observed state, the system detects the update and triggers a new planning cycle. Only the parts of the plan affected by the change are refreshed, while the remaining actions continue without interruption.

\subsection{Module I: Perception-to-Symbolic State}
\label{subsec:p2s3}

\begin{figure}[!t]
    \centering
    \includegraphics[width=\columnwidth]{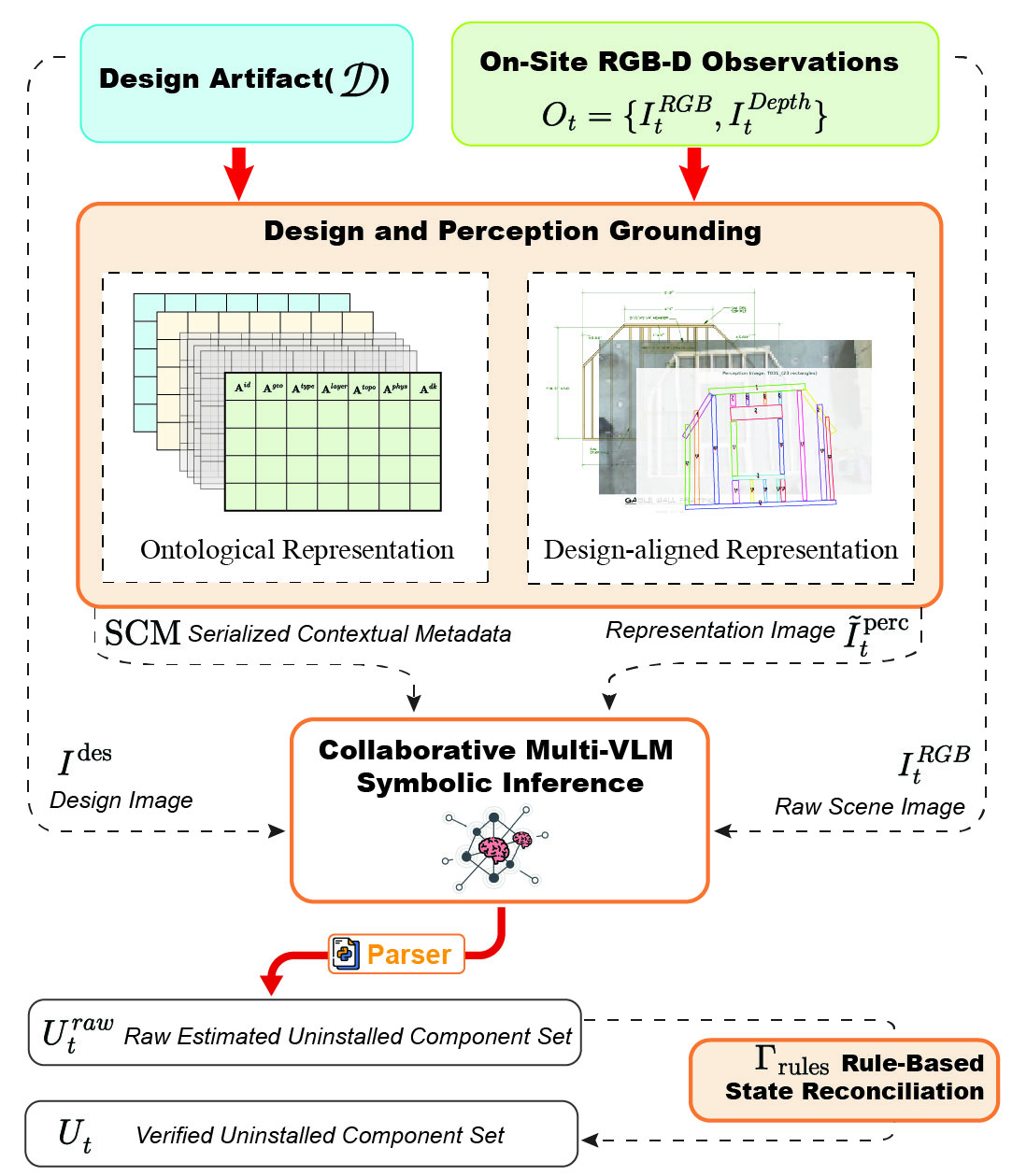} 
    \caption{Overview of the Perception-to-Symbolic State module, which synthesizes a verified symbolic assembly state from design specifications and RGB-D observations.}
    \label{fig:p2s3_diagram}
\end{figure}

As shown in Fig.~\ref{fig:p2s3_diagram}, the PSS module synthesizes verified symbolic assembly states by aligning design specifications with perceptual observations. 
It first converts the design artifact into a structured ontological array $\mathbf{A}$ that encodes component identity, geometry, dependencies, and assembly rules. 
From this representation, a design image $I^{des}$ and serialized contextual metadata (SCM) are generated to provide geometric and semantic priors. 
In parallel, the RGB-D observation $O_t$ is transformed into a design-aligned perceptual image $\tilde{I}_t^{\mathrm{perc}}$.

Given these aligned representations, the multiple VLM-based agents infer a preliminary symbolic hypothesis $U_t^{raw}$ of the uninstalled component set from $(I^{des}, \tilde{I}_t^{\mathrm{perc}})$, the SCM, and the RGB image $I_t^{RGB}$. 
This hypothesis is then refined by a rule-based reconciliation operator $\Gamma_{\mathrm{rules}}$, which enforces the structural, semantic, and geometric constraints encoded in $\mathbf{A}$ to produce a deterministic symbolic state $U_t$. 
The verified state $U_t$ serves as the input to the planning module.

\subsubsection{\textbf{Structured ontological representation}}
\label{subsubsec:sor}

\paragraph{\textbf{The design image $I^{\mathrm{des}}$}}
To initiate this process, the design artifact is first parsed into its geometric elements. Such elements correspond to the basic drawing entities provided by some commonly used design formats (e.g., line and curve segments in CAD file). The system reconstructs component primitives from these inputs.
All structural members are represented in a shared coordinate frame. Using the reconstructed component primitives, the system generates the design image $I^{\mathrm{des}}$, which visualizes the layout of all components in a common coordinate frame. This image defines the reference frame used to align the perceptual representation $\tilde{I}_t^{\mathrm{perc}}$ in the VLMs inference stage.

\paragraph{\textbf{The multidimensional component array $\mathbf{A}$}}

\begin{figure}[!b]
    \centering
    \includegraphics[width=\columnwidth]{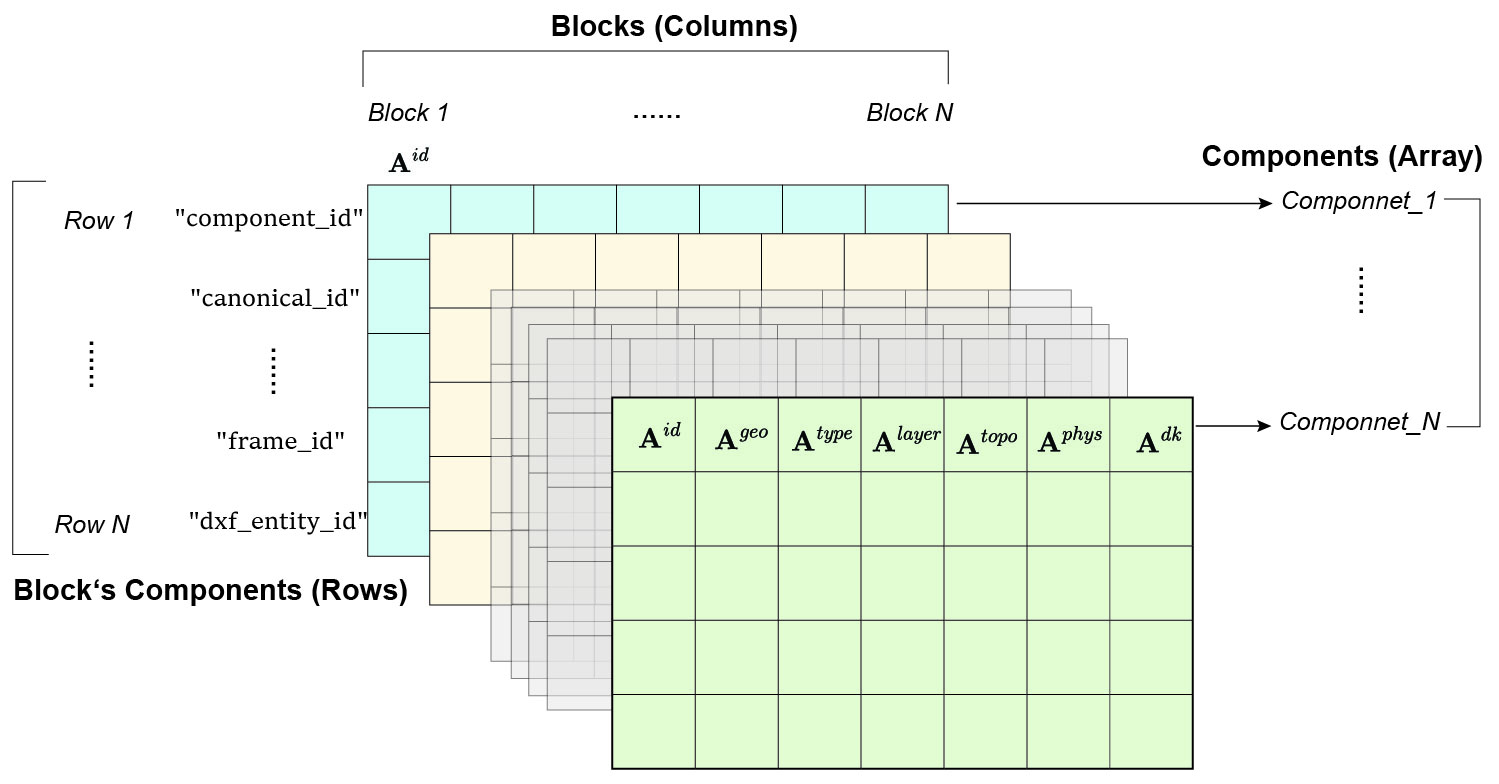}
    \rev{\caption{Structure of the multidimensional component array $\mathbf{A}$.}}
    \label{fig:A_matrix_structure}
\end{figure}

\begin{table*}[!t]
\caption{Definition and functional roles of the component array $\mathbf{A}$.}
\label{tab:A_definition}
\centering
\small
\renewcommand{\arraystretch}{1.25}
\setlength{\tabcolsep}{8pt}

\begin{tabular}{p{0.08\textwidth}|p{0.35\textwidth}|p{0.5\textwidth}}
\toprule
\textbf{Sub-array} & \textbf{Definition} & \textbf{Functional Role} \\
\midrule

$\mathbf{A}^{id}$ &
Defines component identifiers and reference keys. &
Supports indexing across perception and planning. \\
\cmidrule(lr){1-3}

$\mathbf{A}^{geo}$ &
Defines component geometry. &
Supports design alignment and feasibility checking. \\
\cmidrule(lr){1-3}

$\mathbf{A}^{type}$ &
Defines the functional type of each component. &
Constrains interpretation in perception and action selection in planning. \\
\cmidrule(lr){1-3}

$\mathbf{A}^{layer}$ &
Defines layering and grouping relations. &
Supports multi-layer reasoning under occlusion. \\
\cmidrule(lr){1-3}

$\mathbf{A}^{topo}$ &
Defines structural relations. &
Supports dependency checking and admissible frontier computation. \\
\cmidrule(lr){1-3}

$\mathbf{A}^{phys}$ &
Defines physical properties. &
Supports capability-aware task allocation. \\
\cmidrule(lr){1-3}

$\mathbf{A}^{dk}$ &
Defines domain knowledge and assembly rules. &
Guides state reconciliation and enforces valid configurations. \\

\bottomrule
\end{tabular}
\end{table*}


From the same reconstructed primitives, the system constructs the multidimensional component array $\mathbf{A}$, which provides a unified symbolic representation of the design. 
Each row represents a component, and each column group encodes a category of design information. As illustrated in Fig.~\ref{fig:A_matrix_structure}, the array is organized into seven parts:
\[
    \mathbf{A} = \big[\mathbf{A}^{id} \;\big|\; \mathbf{A}^{geo} \;\big|\; \mathbf{A}^{type} \;\big|\; \mathbf{A}^{layer} \;\big|\; \mathbf{A}^{topo} \;\big|\; \mathbf{A}^{phys} \;\big|\; \mathbf{A}^{dk}\big].
\]
Each sub-array captures a specific aspect of component identity, geometry, semantics, structure, or physical feasibility.
Their definitions and functional roles are summarized in Table~\ref{tab:A_definition}.
The instantiation of these sub-arrays is further described in Sec.~\ref{subsec:case_arrays}.

\paragraph{\textbf{Component array to SCM}}

To support the VLMs inference process, $\mathbf{A}$ is converted into the SCM stored in a structured text format such as JSON. The document contains global metadata that records design constants including project identifiers, reference coordinate systems, material standards, and measurement units, and it also contains component-level entries that encode the fields of the seven arrays in $\mathbf{A}$. This serialized metadata serves as the structured textual input for the VLM-based reasoning module.

This ontological representation, together with the rendered design image and the SCM, encodes design geometry and structural dependencies. It defines the geometric frame, the component dependencies, and the rule-based constraints that support reliable symbolic reasoning in next planning stages.

\subsubsection{{\textbf{Design-aligned perceptual representation}}}
\label{subsubsec:dapr}

\begin{figure}[!b]
    \centering
    \includegraphics[width=\columnwidth]{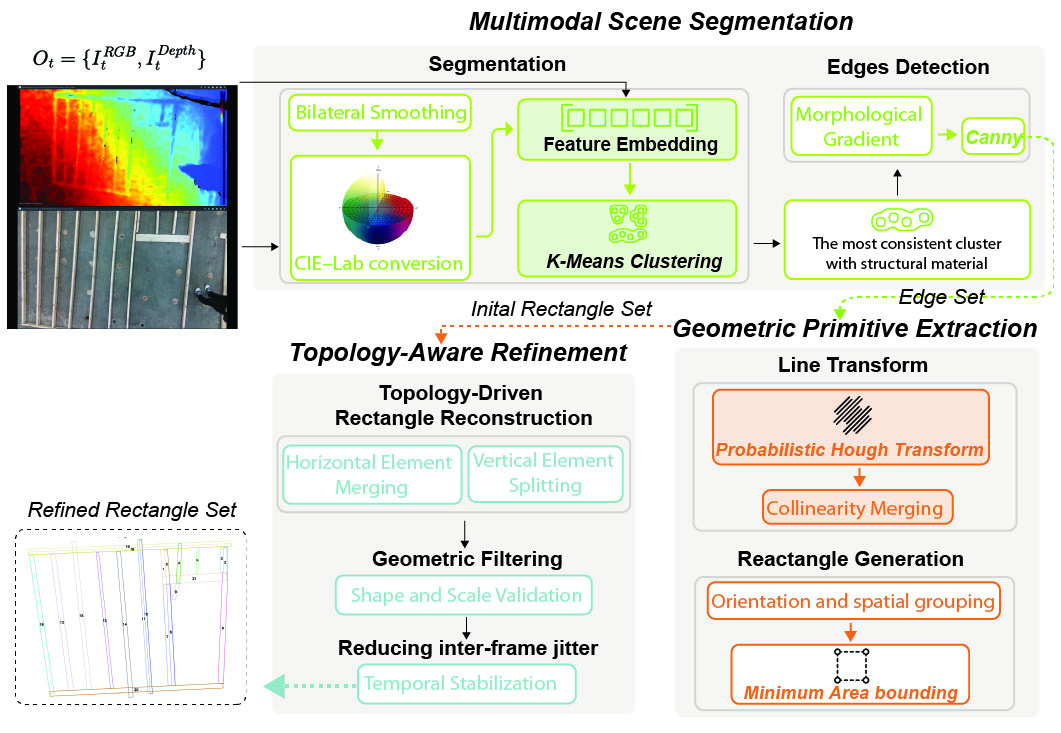}
    \caption{\rev{Design-aligned perceptual representation pipeline for converting raw RGB-D observations into structured geometric primitives.}}
    \label{fig:vision}
\end{figure}

With the design image and the structured component array establishing a common reference frame, the RGB-D observation $O_t$ is transformed into a design-aligned geometric abstraction $\tilde{I}_t^{\mathrm{perc}}$ through multimodal segmentation, primitive extraction, and topology-aware refinement, which is illustrated in Fig.~\ref{fig:vision}.

\paragraph{\textbf{Multimodal scene segmentation}}
The segmentation stage extracts structural regions from the RGB-D observation $O_t$ by suppressing background content.
The RGB image first undergoes background suppression to reduce interference from non-structural objects and illumination variation.
It is then converted into the CIE-Lab color space to improve robustness under varying lighting conditions. Each pixel is embedded into a feature vector that combines its color channels, normalized depth, and normalized spatial coordinates. This multimodal feature space captures photometric and geometric cues relevant to structural material. A clustering procedure partitions the feature space into coherent regions, and the region consistent with expected material appearance and depth characteristics is selected as the candidate assembly region.

Within this region, a morphological gradient operator~\cite{serra1982image} enhances material boundaries, and an edge detector converts these transitions into an edge set that suppresses background clutter. This segmentation result forms a geometric basis for the subsequent primitive extraction stage.

\paragraph{\textbf{Geometric primitive extraction}}
The extracted edge set is transformed into a structured representation of geometric primitives that approximate the visible components. The Probabilistic Hough Transform~\cite{kiryati1991probabilistic} aggregates distributed edge information into a compact set of line hypotheses. Hypotheses that are nearly collinear or overlapping are merged to reduce fragmentation, and the resulting lines are grouped by dominant orientation and spatial proximity. For each group, a minimum-area bounding primitive is estimated to form an initial geometry set. This set reflects specific component shapes and provides a link between raw edges and structural inference.

\paragraph{\textbf{Topology-aware refinement}}
\rev{The initial geometry set is refined through three sequential steps, consistent with the pipeline in Fig.~\ref{fig:vision}, to enforce geometric regularity and structural topology. The refinement process consists of three sequential steps. First, topology-driven rectangle reconstruction merges fragmented primitives that correspond to the same design element. Primitives that span multiple design-defined adjacencies are partitioned at intersection-consistent locations to recover valid component boundaries and relations.
Second, the reconstructed candidates are passed to geometric filtering. This step removes primitives that violate shape and scale plausibility under the design representation and sensing resolution, thereby suppressing false positives caused by clutter, background texture, and depth artifacts.
Finally, temporal stabilization reconciles the current primitive set with the previous frame using local spatial and geometric consistency. This step reduces inter-frame jitter and preserves stable primitive identity. When a vertical candidate becomes abnormally wide relative to its local reference group, it is split to recover a structurally consistent representation under partial occlusion.}

By transforming raw RGB-D observations into a design-aligned perceptual representation, this refinement process yields a structured geometric input suitable for downstream symbolic reasoning.  Implementation details specific to the case study are provided in Section~\ref{subsec:perception}.


\subsubsection{\textbf{VLM-based symbolic inference}}
\label{sec:VLMsmethod}

The VLM-based inference stage compares the design image with the design-aligned perceptual representation to determine which components are installed and which remain uninstalled. To ensure reliable matching, the system uses a structured prompting strategy that provides explicit context descriptions, standardized templates, and a fixed output format. The prompt instructs the model to analyze the detected geometric primitives in $\tilde{I}^{\mathrm{perc}}_t$ and to associate them with the corresponding components in $I^{des}$ under the constraints encoded in the SCM. An additional RGB image $I^{RGB}_t$ is used to support the removal of non-structural detections through a supplementary filtering prompt.

The prompting strategy uses Chain-of-Thought ~\cite{wei2022chain} to structure the reasoning process. The model filters perceptual noise and performs hierarchical component matching under geometric and topological constraints. After the matching geometric assignments are made, the model checks the topological relations to confirm that each component satisfies the required connections and ordering rules. The procedure also applies dependency-based inference, where the model infers missing components when their supporting elements are detected and no conflicting evidence is present. This stage's output is the initial uninstalled set $U_t^{raw}$, which contains component-level match decisions and possible inference labels. This set is then refined by a deterministic reconciliation operator that enforces structural constraints to correct missing assignments and remove inconsistent matches.


\subsubsection{\textbf{Rule-based state reconciliation}}
\label{subsec:rule_reconciliation}
The reconciliation operator $\Gamma_{\mathrm{rules}}$ refines the initial perceptual hypothesis $U_t^{raw}$ into a verified symbolic state using a finite set of atomic rules. \revtwo{Each assembly constraint rule is defined as:
\begin{equation}
\mathbf{A}^{dk}_{k,i} = \{ (v_j, v) \mid v_j \in V, v \in \{0,1\} \}.
\label{eq:dk_form}
\end{equation}
Here, $v=1$ specifies a prerequisite component $v_j$ that must be present in the state, while $v=0$ imposes an exclusion constraint. Based on these conditions, $\Gamma_{\mathrm{rules}}$ executes three functional rule types: \textit{Completion} rules assert the presence of components when all required predecessors are identified; \textit{Pruning} rules eliminate components from the hypothesis that violate prerequisite or exclusion constraints; and \textit{Refinement} rules evaluate composite dependencies to resolve state ambiguities. 
By applying the rule sequence $\mathbf{A}^{dk}_1,\ldots,\mathbf{A}^{dk}_m$ in a fixed order starting from $U_t^{raw}$, $\Gamma_{\mathrm{rules}}$ overrides conflicting updates to converge on a deterministic symbolic state.} 

\revtwo{The number, ordering, and instantiation of atomic rules depend on the topological complexity of the design and are reported for the timber-frame case study in Sec.~\ref{subsec:VLMs_case}.}

\begin{algorithm}[!t]
\small
\caption{Reactive Closed-Loop Planning and Execution}
\label{alg:hpr2_main_new}
\begin{algorithmic}[1]
\REQUIRE design artifact $\mathcal{D}$, component array $\mathbf{A}$, robot team $\mathcal{R}$
\STATE $t \leftarrow 0$; $P_{-1} \leftarrow \varnothing$; $U_{-1} \leftarrow V$

\WHILE{$U_{t-1} \neq \varnothing$}

    \STATE \textbf{State Update}
    \STATE $O_t \gets \textsc{Perceive}(t)$
    \STATE $U_t^{raw} \gets PSS(\mathcal{D}, \mathbf{A}, O_t)$
    \STATE $U_t \gets \Gamma_{\mathrm{rules}}(U_t^{raw}, \mathcal{D}, \mathbf{A})$
    \STATE $I_t \gets V \setminus U_t$
    \STATE $\mathcal{E}_t \gets \textsc{DetectEvent}(U_{t-1}, U_t, P_{t-1})$

    \STATE \textbf{Frontier Evaluation}
    \STATE $T_t \gets \textsc{Frontier}(U_t, \mathcal{D})$

    \STATE \textbf{Plan Update}
    \IF{$\mathcal{E}_t = \textsc{None}$}
        \STATE $P_t \gets P_{t-1}$
    \ELSE
        \STATE Preserve valid assignments from $P_{t-1}$ for tasks in $T_t$
        \STATE $P_t \gets$ assign remaining admissible tasks in $T_t$ to $\mathcal{R}$
    \ENDIF
    
    \STATE \textbf{BT Update and Execution}
    \FOR{each $r_k \in \mathcal{R}$}
        \STATE $\mathcal{B}_t^{k} \gets \textsc{BT\_Update}(\mathcal{B}_{t-1}^{k}, P_{t-1}, P_t)$
    \ENDFOR
    \STATE \textsc{TickBT\_Executor}($\{\mathcal{B}_t^{k}\}$)

    \STATE $t \gets t+1$

\ENDWHILE

\RETURN $P_{t-1}$
\end{algorithmic}
\end{algorithm}

\subsection{Module II: Human-Aware Planning and Replanning}
\label{sec:module2}

The HPR module updates and executes task assignments within the closed-loop procedure in Algorithm~\ref{alg:hpr2_main_new}. 
At each iteration $t$, it computes the admissible frontier $T_t$ from the current symbolic state $(I_t,U_t)$, updates the task assignment plan $P_t$ by preserving unchanged assignments and allocating remaining admissible tasks under feasibility constraints, and updates per-robot behavior trees $\mathcal{B}_t^{k}$ for execution.

\subsubsection{\textbf{Human-aware replanning}}
\rev{Following the definition of human awareness introduced in Sec.~I, human intervention here is modeled as a replanning event $\mathcal{E}_t$.}
It is detected by comparing the updated uninstalled set $U_t$ with the expected outcome of the previous robot execution.
If no event is detected, the existing plan is retained.
Otherwise, the planner selects a new feasible assignment that minimally deviates from the previous one, according to the minimal-change rule in~\eqref{eq:minchange} and the plan-edit distance in~\eqref{eq:edit_distance}. The resulting assignment is executed by the robots, and execution outcomes are fed back to the loop until all components are installed.

\subsubsection{\textbf{Frontier evaluation and feasibility assessment}}

Given the current uninstalled set $U_t$, the planner identifies the admissible frontier $T_t$ of components whose design prerequisites are satisfied.
For each frontier component, the planner evaluates feasible robot assignments under the current observation.
Feasibility is determined by the robot capability profiles $C_j$, including payload limits, kinematic reach, collision safety, and perceived human occupancy regions.

\subsubsection{\textbf{Minimal-change task reassignment}}

Replanning is triggered only when a human-induced or unexpected state change is detected.
When replanning is required, the planner updates the task assignment by approximately minimizing the plan-edit distance $d(P_t,P_{t-1})$.
Assignments from the previous planning cycle associated with components that remain admissible under the updated symbolic state are retained.
Admissibility here requires that structural prerequisites remain satisfied despite human intervention, which avoids unnecessary increases in the plan-edit distance.
Components that are not preserved from the previous assignment are selected for reassignment.
This set includes components that become newly admissible due to state changes and components whose prior assignments are no longer feasible.
\rev{The preservation and reassignment check is performed through a single pass over the overlapping frontier, giving the minimal-change update a linear worst-case time complexity of $\mathcal{O}(|T_t|)$.}
For these components, the planner performs task allocation using a workload-based heuristic while enforcing the defined feasibility constraints.

\subsubsection{\textbf{Behavior tree generation and update}}

The updated task assignment is projected onto robot-specific behavior trees.
For each robot $r_k$, execution is represented by a behavior tree $\mathcal{B}_t^{k}$.
Tasks whose assignments are unchanged preserve their corresponding subtrees in $\mathcal{B}_{t-1}^{k}$.
Only reassigned components generate new or modified subtrees in $\mathcal{B}_t^{k}$.
Single-robot tasks are mapped to sequential execution structures.
Tasks requiring multiple robots include synchronization nodes that enforce coordinated execution across the corresponding trees.
Behavior trees are executed asynchronously to allow parallel execution.


\section{Case Study}
\label{sec:case}

\begin{figure}[!t]
 \centering
 \includegraphics[width=\columnwidth]{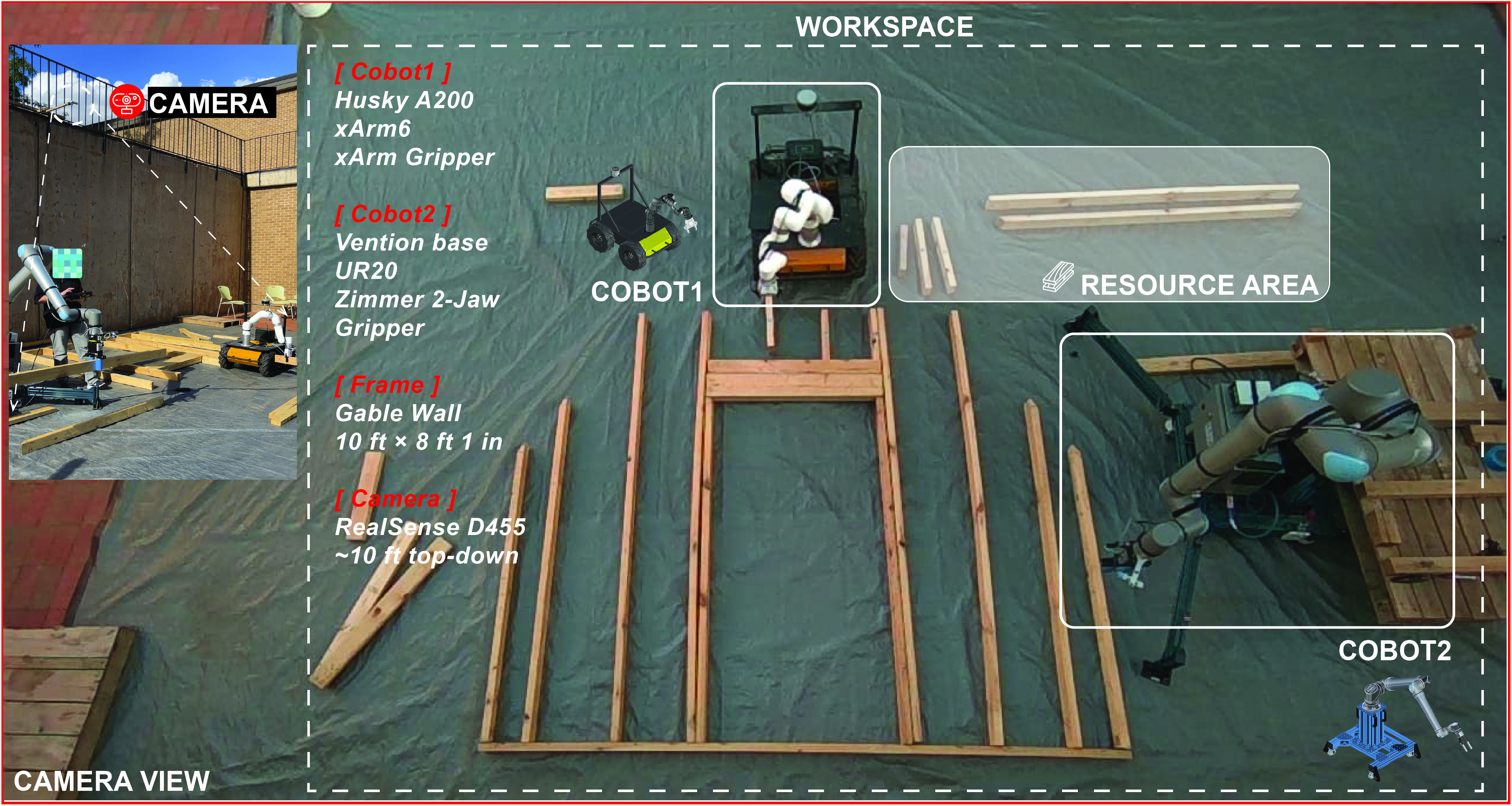}
 \caption{Example layout for the case study. The configuration includes two heterogeneous collaborative robots, an overhead RGB-D camera, and the design coordinate frame registered to the workspace.}
 \label{fig:system_overview}
\end{figure}

The timber-frame wall is a representative structured assembly task with geometric similarity across members and strict prerequisite relations. These properties lead to ambiguous matches under clutter and occlusion. The frame also contains components with different physical properties and functional roles. These include short studs that can be handled by a single robot and long plates that require cooperative handling. Human workers may install or remove any component at unscripted times. These actions introduce sudden changes in the uninstalled set, alter feasible coalitions, and create conflicts with ongoing robot tasks. The combination of these factors creates a realistic environment that supports a comprehensive evaluation of the robustness of our method.

\rev{Figure~\ref{fig:system_overview} shows the HRC timber-assembly workcell used in the case study, including two heterogeneous collaborative robots, an overhead RGB-D camera with a registered top-down view, the resource area, and the installation workspace under the shared design coordinate frame.} \revtwo{This workcell provides the layout used for the real-site RGB-D data in Sec.~\ref{subsec:perception}; the planning experiments in Sec.~\ref{subsec:case_planning} uses the same layout and robot specifications in task-level simulation.}

\subsection{\rev{Experimental design for real-site validation}}
\label{subsec:exp_design_pss}

To enable a controlled evaluation under real-site conditions, we design a grouping strategy that isolates the effects of varying sensing conditions and structural layouts. Because the perception module relies primarily on geometric abstractions and visual contrast rather than specific material textures, variations in material type do not alter the perception-to-symbolic mapping as long as sufficient color contrast is maintained between the target elements and the background. Consequently, rather than introducing additional material variations, our evaluation prioritizes structural diversity to assess the framework's generalization across assembly tasks with distinct topological characteristics and spatial scales.

\begin{figure*}[!h]
    \centering
    \includegraphics[width=\textwidth]{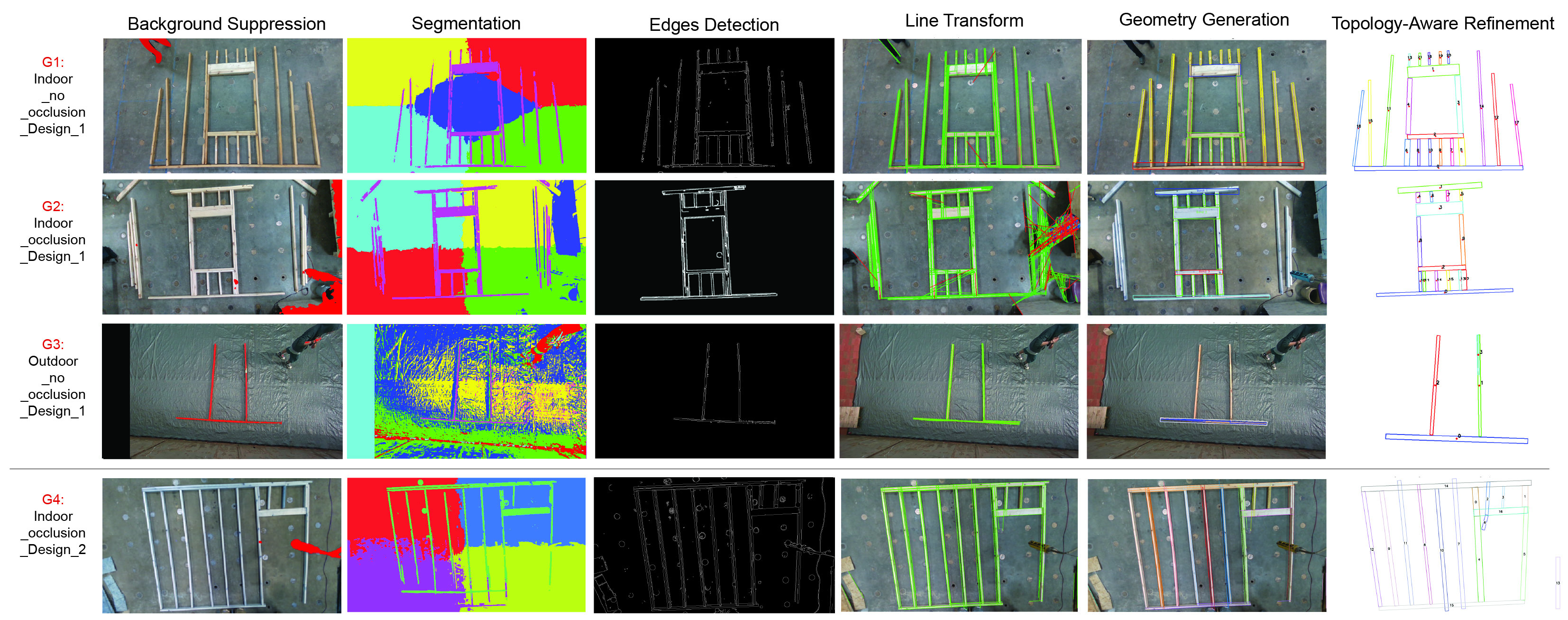}
    \caption{\rev{Design-aligned perceptual representation pipeline results across four evaluation groups (G1-G4).}}
    \label{fig:G4}
\end{figure*}

To this end, we adopt two representative design configurations. The baseline configuration is a symmetric \emph{gable-wall} layout comprising 27 geometric primitives with overall dimensions of approximately $10 \times 8.1$ ft, as shown on the right side of Fig.~\ref{fig:ontology_pipeline}. To introduce structural variation, we further include an asymmetric \emph{stair-wall} layout consisting of 17 geometric primitives with dimensions of approximately $11.8 \times 9.1$ ft, as illustrated in Fig.~\ref{fig:G4}. These configurations provide complementary test cases to evaluate the robustness of the pipeline against variations in topology, scale, and spatial arrangement.

\rev{To instantiate these configurations under controlled sensing conditions, we construct four evaluation groups (G1-G4) to decouple different sources of variation. \textbf{G1} (Indoor, Gable-wall, Uncluttered) represents a baseline setting with minimal occlusion and clutter, serving as a low-noise reference. \textbf{G2} (Indoor, Gable-wall, Cluttered) retains the gable-wall layout but introduces human presence, partial occlusion, and on-site clutter to capture perceptual interference. \textbf{G3} (Outdoor, Gable-wall, Uncluttered) preserves the same structural layout as G1 but shifts the setup to an outdoor environment, emphasizing changes in illumination and background conditions. \textbf{G4} (Indoor, Stair-wall, Cluttered) adopts the asymmetric stair-wall layout under indoor clutter conditions comparable to G2, supporting evaluation of structural variation under similar perceptual interference. Together, these four groups establish a controlled basis for evaluating the robustness and cross-layout generalization of the proposed framework.}
\subsection{Ontological representation of the DXF design file}
\label{subsec:case_arrays}

\begin{figure}[!b]
    \centering
    \includegraphics[width=\columnwidth]{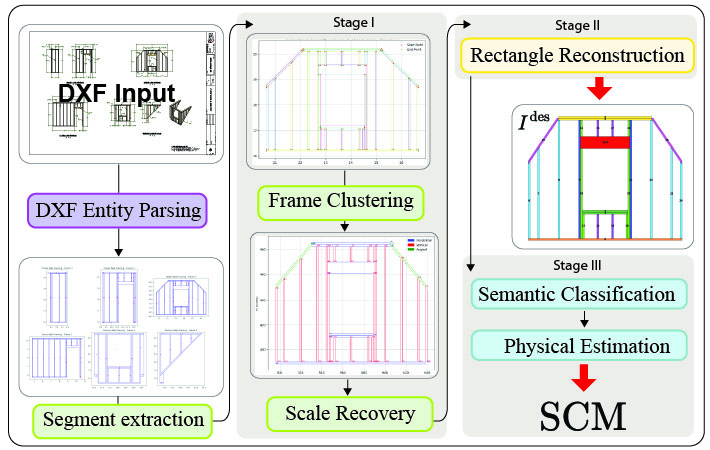}
    \caption{\rev{Pipeline for generating the ontological representation.}}
    \label{fig:ontology_pipeline}
\end{figure}

The elevation drawing of this wall is available in Drawing Exchange Format (DXF) format and serves as the design artifact for constructing the ontological representation. The processing workflow, illustrated in Fig.~\ref{fig:ontology_pipeline}, begins by decomposing the DXF file into its elementary CAD entities, including \texttt{LINE} and \texttt{LWPOLYLINE}. Each polyline is subdivided into its constituent segments, producing a detailed geometric description of the drawing. These segments form the primitive units from which the structural layout is recovered.
To recover the structural layout, segments are grouped into coherent component clusters based on spatial adjacency, supported by a reading-order prior that follows the typical top-to-bottom, left-to-right organization of architectural drawings.

To ensure metric consistency, we recover scale by mapping DXF units to physical dimensions, assigning the shortest detected segment to the nominal lumber thickness of 1.5 inches.
All geometric quantities are then expressed in real-world units.
Rectangular component candidates are reconstructed by pairing parallel and nearly equal-length line segments separated by referring to the nominal lumber thickness. This procedure yields the horizontal, vertical, and diagonal rectangles that form the geometric footprint of the timber members. 

\begin{figure}[!b]
    \centering
    \includegraphics[width=\columnwidth]{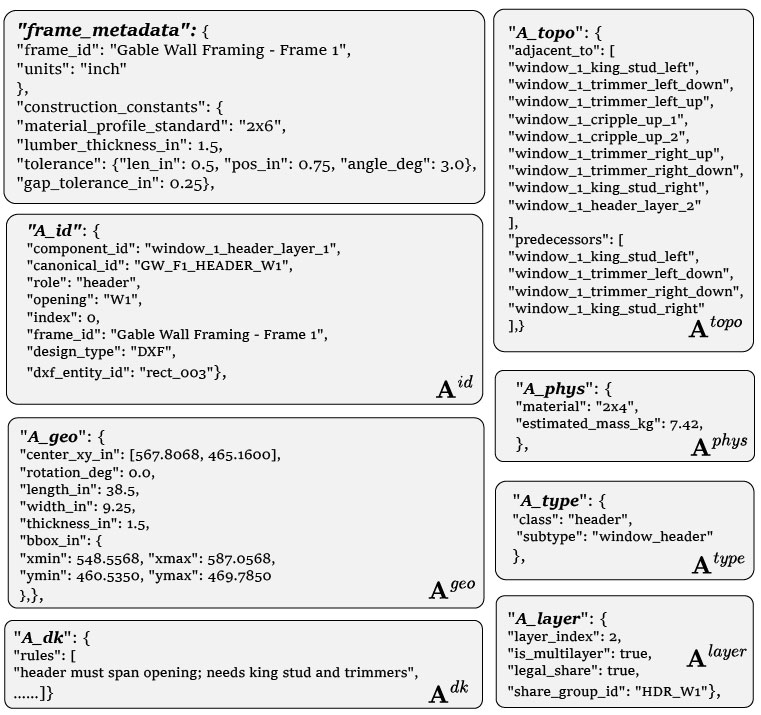}
    \caption{Representative example of the arrays in $\mathbf{A}$ for the component \texttt{window\_1\_header\_layer\_1}.}
    \label{fig:A_example}
\end{figure}

Using these reconstructed rectangles, the design image $I^{\mathrm{des}}$ is rendered in the unified coordinate frame, where each rectangle is drawn with its detected geometry and assigned a fixed color that denotes its construction role (e.g., plates in yellow, headers in red, studs in cyan). These color encodings serve as semantic cues that guide VLM-based matching by disambiguating component types during visual reasoning.
The resulting image provides the geometric and spatial reference used during VLMs reasoning.
The same rectangles are used to instantiate the $\mathbf{A}$, from which the SCM file is serialized in JSON format. 
Fig.~\ref{fig:A_example} illustrates the global metadata and the seven array fields for a representative component (\texttt{window\_1\_header\_layer\_1}).

This example demonstrates how the abstract representation introduced in Sec.~\ref{subsubsec:sor} is instantiated for a real assembly task. 
In this example, the identity array $\mathbf{A}^{id}$ specifies the component identifier, canonical name, opening association, and the originating DXF entity. The geometry array $\mathbf{A}^{geo}$ explicitly records rectangle pose and dimensions, including its center coordinates, length, width, orientation, and bounding box in design units.

Semantic role information is instantiated in $\mathbf{A}^{type}$ by labeling the component as a window header, while $\mathbf{A}^{layer}$ indicates its membership in a multi-layer group through a shared group identifier. The topological array $\mathbf{A}^{topo}$ lists adjacent and predecessor components. Physical properties such as the estimated mass are stored in $\mathbf{A}^{phys}$ and later used for capability-aware task allocation.
Finally, $\mathbf{A}^{dk}$ encodes construction-specific assembly rules that enable deterministic reconciliation under ambiguous perceptual observations.

\subsection{Design-aligned representation of real-site RGB-D images}
\label{subsec:perception}

\rev{The real-site RGB-D data used in this section are collected under the four controlled groups (G1--G4) defined in Sec.~\ref{subsec:exp_design_pss}. Indoor data were captured using an Intel RealSense D435, while outdoor data used a RealSense D455, whose global shutter improves depth stability under strong illumination. Both sensors operate at a resolution of $1280 \times 720$ with built-in intrinsic parameters for metric projection. In total, the dataset comprises 400 RGB-D instances, with 100 samples per group. Example images are shown in the first column of Fig.~\ref{fig:G4}. The experiments use a single overhead RGB-D camera as a controlled sensing configuration to evaluate whether the proposed PSS module can operate under partial visibility, clutter, and environmental variation using a basic sensing setup. In practical deployment, multi-view sensing may be needed when the assembly area is large, when components are frequently occluded, or when geometric primitives cannot be observed from a single viewpoint.}

Following the procedure outlined in Sec.\,\ref{subsubsec:dapr}, the RGB-D observations are conditioned to extract structural regions and suppress background interference. Although the segmentation follows a common sequence, the relative weighting of appearance and depth cues differs between indoor and outdoor settings due to their sensing characteristics.
For indoor scenes, depth is stable under uniform lighting, but clutter and floor texture introduce background edges.
We therefore apply background suppression and hole filling before clustering. Background regions are removed, small distractions are filled, and each pixel is embedded in a six-dimensional feature vector
\[
[L, a, b,\; w_x x,\; w_y y,\; w_z z],
\]
where $(L,a,b)$ denote pixel color in the CIELAB color space, $(x,y)$ are normalized image coordinates, and $z$ is normalized depth.
The weights $w_x$, $w_y$, and $w_z$ balance appearance, spatial proximity, and depth cues.
They are empirically selected per setting and kept fixed across all experiments.
The K-means clustering~\cite{macqueen1967kmeans} algorithm is used to group the pixels into coherent regions, and a scoring rule selects the cluster that best matches the expected structural geometry by scoring from color, depth, and the pixels coordination. For outdoor scenes, the segmentation assigns a higher weight to depth cues than to appearance cues. Pixels with inconsistent depth are filtered out, and the resulting depth mask restricts the clustering stage to regions more likely to contain structural elements.

After segmentation, a morphological gradient~\cite{serra1982image} highlights the component boundaries, and the Canny detector~\cite{canny1986computational} converts these boundaries into edge maps used for geometric processing.
Fig.~\ref{fig:G4} shows representative results for both indoor and outdoor settings, including masked RGB images after the background suppression step, the segmentation mask result, and the raw Canny edges. The filtered edges are converted into geometric elements in this stage. Local edge patterns are aggregated into line segments using the Probabilistic Hough Transform (PHT)~\cite{kiryati1991probabilistic}.  
Segments that lie along nearly the same direction are merged, then grouped according to their main orientation and location.  
Because timber-frame structures are rectilinear, each group is converted to a rectangle based on the Freeman and Shapira~\cite{freeman1991minrect}'s minimum-area bounding theory.  This produces an initial set of rectangles that outline candidate structural components.

The initial rectangles are further adjusted to align with the structure of the wall frame. Long horizontal members are merged if they appear fragmented. Vertical elements that cross these horizontals are divided at the intersection to recover the correct structural layout.  Rectangles that are clearly too stretched, too short, or inconsistent in scale are removed through common geometric checks. Finally, each rectangle is compared with its counterpart in the previous frames, which reduces flicker caused by depth noise, glare, or temporary occlusions from people passing through the scene. 
Fig.~\ref{fig:G4} shows representative outputs, including lines set, geometric groupings, and the final rectangle set.

The final output is a clean set of design-aligned rectangles that summarizes the visible structure. This representation is then passed to the VLMs reasoning stage with other outputs in the ontological representation procedure. \rev{For reproducibility, core hyperparameter configurations for the PSS module are summarized in Appendix~\ref{app:pss_reproducibility}.}

\subsection{VLM-based semantic inference in real scenes}
\label{subsec:VLMs_case}

\begin{figure*}[!h]
    \centering
    \includegraphics[width=\textwidth]{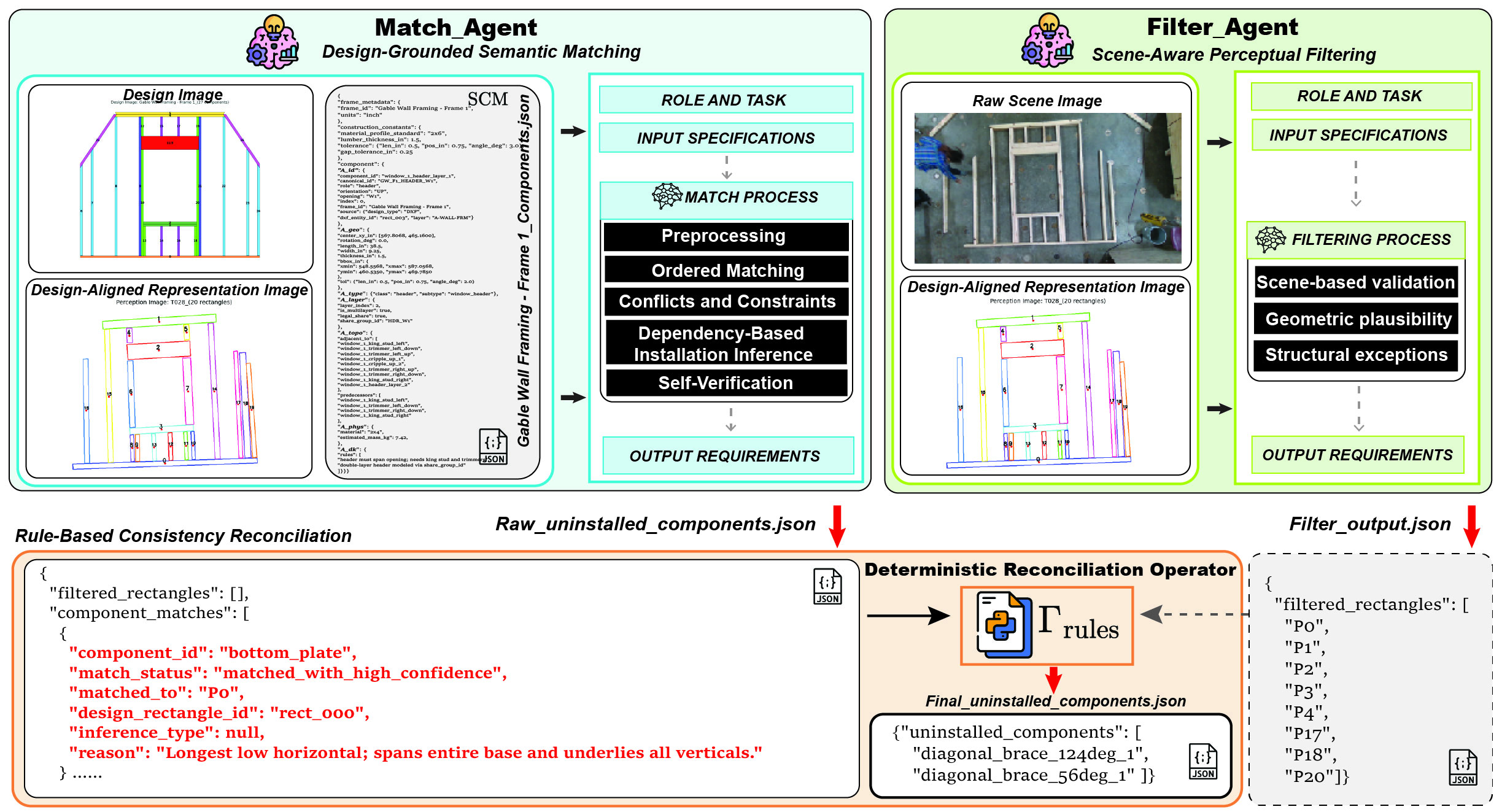}
    \caption{VLM-based symbolic inference workflow, illustrating the interaction between the Match Agent, the Filter Agent, and rule-based consistency reconciliation.}
    \label{fig:VLMs_cot}
\end{figure*}

Fig.~\ref{fig:VLMs_cot} shows the VLM-based semantic inference workflow used in the case study.
The workflow employs two VLM-based reasoning agents that operate on the same design-aligned perceptual representation.
A Match Agent performs semantic inference grounded in the design specification by aligning perceptual primitives with the design ground truth.
In parallel, a Filter Agent evaluates scene consistency by identifying perceptual primitives that are geometrically or visually inconsistent with the observed scene.
The outputs of both agents are jointly evaluated by a deterministic reconciliation operator, which produces the final verified uninstalled set. The following paragraphs describe how this process operates under real-site sensing conditions. \rev{And the prompts used to guide the Match Agent and the Filter Agent are detailed in Appendix~\ref{app:promptA} and Appendix~\ref{app:promptB}.} 

The Match Agent receives the design image, the design-aligned perceptual representation, and the SCM in a shared coordinate frame. The agent is guided through a series of reasoning steps aligned with construction knowledge. The matching procedure begins with elements that are easiest to locate. For example, long horizontal members are identified first because they outline the wall and set clear positional limits for other parts. Once these boundaries are recognized, the model determines the geometry of the opening, including its sill, header, and surrounding spans. The agent then resolves the major vertical components near the opening by checking their layout relative to the established boundaries. Studs across the frame are interpreted next with attention to spacing, alignment, and the type rules encoded in the SCM. Diagonal elements are resolved last by evaluating their orientation and attachment relations. Finally, a verification step compares the inferred configuration with structured constraints. 

For the Filter Agent, it evaluates perceptual consistency by removing geometric primitives that do not correspond to visible structural members.
It operates on the same design-aligned perceptual representation and RGB image.
In uncluttered scenes, this process preserves the original perceptual set.
In cluttered scenes, it suppresses spurious detections caused by background objects, temporary occlusions, or depth artifacts.

In practice, both agents output their reasoning results in a structured JSON format. The Match Agent reports, for each design component, its predicted installation status, matched perceptual primitive, and a brief justification.
The Filter Agent returns the subset of perceptual primitives consistent with the observed scene.
\revtwo{To combine these outputs into a verified uninstalled set, the reconciliation operator is instantiated for this timber-frame case study with $m=10$ ordered constraint rules that evaluate a state space of $N=27$ components. By enforcing the structural and topological requirements of the design, these rules resolve inconsistencies in the raw semantic hypothesis $U_t^{raw}$ caused by visual occlusion, clutter, or VLM reasoning errors. 
For instance, short vertical elements near window openings omitted due to partial occlusion are reinstated. Inferred horizontal members lacking requisite supporting studs, as well as isolated vertical studs hallucinated in cluttered areas, are eliminated. The presence of critical components (e.g., primary window headers) is asserted only when all structural prerequisites are verified and conflicting layers are absent. Applying this sequence of rules yields a verified, topologically consistent symbolic state.}

\subsection{Planning instantiation for timber frame task}
\label{subsec:case_planning}

Following the closed-loop planning process described in Sec.~\ref{sec:module2}, we instantiate the planner by grounding frontier evaluation and capability constraints in the component arrays.

\rev{The reported planning experiment focuses on the task-planning level. In this case study, task feasibility is evaluated using payload capacity, while kinematic reachability, continuous trajectory feasibility, and collision checking are treated as lower-level motion planning and safety-layer constraints.}
\rev{This setting assumes a sufficiently open workspace with operational separation; therefore, geometric human-robot conflicts are not explicitly evaluated in the current task-level planning experiment.}
Component masses are obtained from the physical attribute array $\mathbf{A}^{phys}$.
A robot team is feasible for a component if its combined payload capacity exceeds the component mass.
\rev{Human interventions are represented as discrete installation or removal state updates, and replanning is triggered by symbolic-state inconsistency events.}
\revtwo{These assumptions reflect a separation between what the task-planning layer determines and what the underlying motion and safety layers execute: the framework decides what should be installed next, by which robot, and how to respond to a detected state inconsistency, while trajectory generation, grasping, and physical safety enforcement are delegated to independent, lower-level mechanisms. This separation does not affect the validity of the contributions evaluated in this case study, since the reported metrics, namely state synthesis accuracy, plan edit distance, and workload balance, are defined entirely at the level of the symbolic state and task assignment and do not depend on how the motion and safety layers are instantiated.}

\begin{figure*}[!h]
  \centering
  \includegraphics[width=\textwidth]{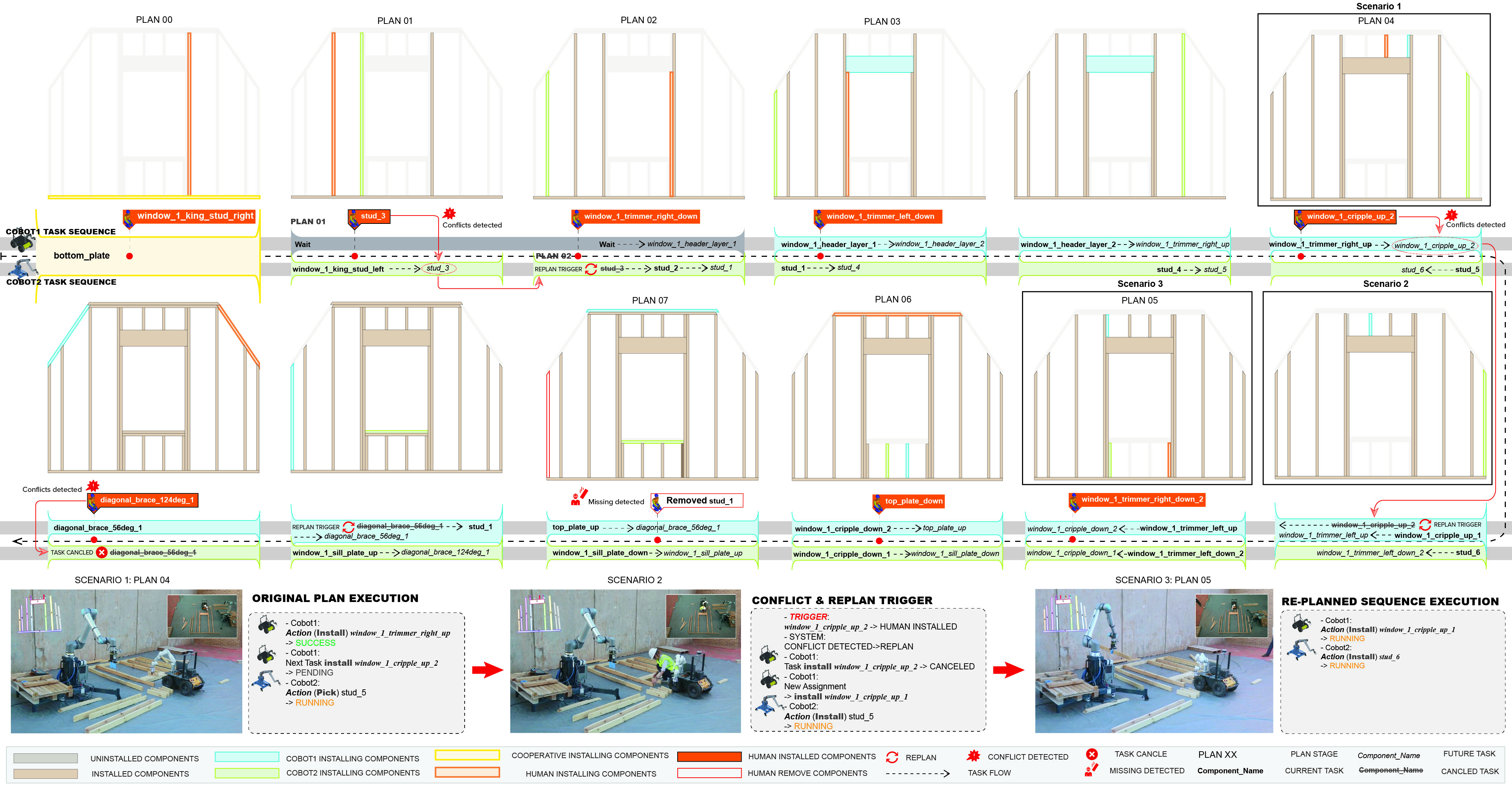}
  \caption{Closed-loop execution trace of the full timber-frame wall assembly, illustrating task progression and replanning events.} 
  \label{fig:assembly_progress}
\end{figure*}

The admissible frontier is derived from prerequisite relations encoded in the design knowledge array $\mathbf{A}^{dk}$.
For example, bottom plates must be installed before studs.
Studs adjacent to openings must be installed before headers.
Components above an opening become admissible only after their supporting studs are complete.
The frontier order produced during planning follows directly from these domain rules.

Heterogeneous payload capacities, together with the minimal-change update, allow task assignments to remain consistent across planning cycles.
Robots that remain feasible for a component retain their assignments.
Components that become newly admissible are assigned to feasible robots with lower current workload.
Human interventions are modeled as installing or removing components.
Such actions modify the uninstalled set and may introduce newly admissible components or invalidate existing assignments.
In these cases, only the affected components are reassigned, while all other assignments remain unchanged.

The resulting coalitions are directly mapped to robot behavior tree execution structures.
For cooperative coalitions, each robot receives a parallel branch containing a shared synchronization node.
Robots execute their \textsc{Move} and \textsc{Co-Pick} actions independently and wait at the synchronization point until all coalition members are ready.
The group then proceeds to a joint \textsc{Co-Install} action, ensuring coordinated placement of heavier components.
Single-robot coalitions yield linear \textsc{Move}-\textsc{Pick}-\textsc{Install} subtrees without synchronization.
Robots remain idle when no component in the current frontier is assigned to them.

Changes in component admissibility and task reassignment under human intervention are visualized in the execution trace in Fig.~\ref{fig:assembly_progress}. For example, the replanning scenario shown in the lower-right panel of Fig.~\ref{fig:assembly_progress} demonstrates a human-triggered plan update during execution.
In the original plan execution, the planner assigns Cobot~1 to install \texttt{window\_1\_trimmer\_right\_up}, while Cobot~2 executes a pick action for \texttt{stud\_5}, both of which are admissible at that iteration. In the next iteration, a human installs \texttt{window\_1\_cripple\_up\_2} ahead of the planned sequence.
This action introduces a discrepancy between the expected and observed installation states, which is detected by the state update module and triggers a replanning event. Following the minimal-change update rule, only the assignments affected by this discrepancy are revised.
The task associated with \texttt{window\_1\_cripple\_up\_2} is canceled, while unaffected assignments are preserved. In the re-planned execution, Cobot~1 is reassigned to install \texttt{window\_1\_cripple\_up\_1}, whereas Cobot~2 continues its previously assigned task without interruption.

\begin{table}[!b]
\centering
\caption{\rev{Mean perceptual complexity descriptors by group (G1-G4).}}
\label{tab:group_descriptor_means}
\setlength{\tabcolsep}{3 pt}
\renewcommand{\arraystretch}{1.05}
{\small
\begin{tabular}{lllccc}
\toprule
\textbf{}
& \makecell{\textbf{condition}}
& \makecell{\textbf{design}}
& \makecell{\textbf{Appearance}\\\textbf{confounders}}
& \makecell{\textbf{Peripheral}\\\textbf{clutter}}
& \makecell{\textbf{Human}\\\textbf{occlusion}} \\
\midrule
G1 & Indoor  & Gable-wall      & 0.16 & 0.42 & 10\% \\
G2 & Indoor  & Gable-wall      & 6.18 & 7.27 & 29\% \\
G3 & Outdoor & Gable-wall      & 0.33 & 2.30 & 3\% \\
G4 & Indoor  & Stair-wall      & 3.34 & 3.55 & 15\% \\
\bottomrule
\end{tabular}
}
\end{table}

\subsection{Quantitative evaluation and results}
\label{subsec:validation}
\begin{table*}[!t]
\centering
\caption{\rev{VLM symbolic matching correct rate by group and inference stage (\%).}}
\label{tab:VLMs_by_group}
\setlength{\tabcolsep}{4pt}
\renewcommand{\arraystretch}{1.05}
{\small
\begin{tabular}{@{}l ccc ccc ccc ccc@{}}
\toprule
\textbf{Model}
& \multicolumn{3}{c}{\textbf{G1}}
& \multicolumn{3}{c}{\textbf{G2}}
& \multicolumn{3}{c}{\textbf{G3}}
& \multicolumn{3}{c}{\textbf{G4}} \\
\cmidrule(lr){2-4} \cmidrule(lr){5-7} \cmidrule(lr){8-10} \cmidrule(lr){11-13}
& Base & Grounded & Verified
& Base & Grounded & Verified
& Base & Grounded & Verified
& Base & Grounded & Verified \\
\midrule
GPT-5
& 52\% & \textbf{95\%} & \textbf{95\% }
& 51\% & \textbf{89\%} & \textbf{98\%$\uparrow$}
& 49\% & \textbf{83\%} & \textbf{83\%}
& 60\% & 74\% & 81\% \\
Gemini-3-Pro
& 42\% & 82\% & 82\%
& 46\% & 74\% & 83\%
& \textbf{61\%} & \textbf{48\%$\downarrow$} & 48\%
& 51\% & \textbf{80\%} & \textbf{85\%$\uparrow$} \\
Claude-Opus-4.5
& 24\% & 71\% & 71\%
& 21\% & 40\% & 85\%
& 28\% & 34\% & 34\%
& 26\% & 55\% & 64\% \\
\bottomrule
\end{tabular}
}
\vspace{0.25em}
\end{table*}
This section evaluates the two core components of the proposed framework.
The first experiment assesses whether the PSS module can convert real RGB-D observations into symbolic states suitable for task planning.
The second experiment evaluates the planning system in a task-level planning simulation without physical execution, focusing on payload-aware allocation and plan adaptation under human intervention.
Detailed experimental settings and results are reported in the following subsections.

\subsubsection{VLMs semantic inference evaluation}
\label{subsubsec:VLMs_eval}

\rev{To enable fine-grained quantitative analysis beyond coarse group labels, we introduce four perceptual complexity descriptors that quantify perceptual complexity and interference. The descriptors include: (i)~\emph{scene complexity}, measured by the average number of target frame components per image; (ii)~\emph{appearance confounders}, measured by the average number of visually confusable non-target instances per image, such as objects with similar hue or elongated shapes; (iii)~\emph{peripheral clutter}, measured by the average number of co-present site elements per image, such as tools, equipment, and people adjacent to the work area; and (iv)~\emph{human occlusion of the frame}, measured as the percentage of images in which human occlusion is present.}

\rev{Table~\ref{tab:group_descriptor_means} summarizes the composition of the dataset and the distribution of perceptual difficulty factors across groups. The count-based descriptors are reported as the average number of annotated instances per image within each evaluation group. The results confirm that G1 serves as a low-noise baseline, G2 introduces high clutter and occlusion, G3 isolates environmental variation, and G4 combines structural variation with moderate perceptual complexity.}

\rev{To evaluate the proposed framework, we first conducted a preliminary screening of several representative VLMs, including both commercial models and open-source alternatives. The tested open-source models achieved limited component-level accuracy, with correct predictions mainly on simple single-component cases, and produced incomplete structured outputs, unsupported component identifiers, or JSON that did not follow the required schema for more complete assembly states. The main quantitative evaluation therefore focuses on models that reliably supported valid structured output.} \rev{Based on this screening, we selected GPT-5, Gemini-3-Pro, and Claude-Opus-4.5 to evaluate the PSS module under three input configurations. \textit{Base} compares the raw RGB scene image with the design rendering. \textit{Grounded} applies design-grounded semantic matching. \textit{Verified} augments grounded matching with a consistency-filtering agent and deterministic reconciliation. The resulting accuracy across models, groups, and inference configurations is summarized in Table~\ref{tab:VLMs_by_group}.}

\begin{figure}[!b]
  \centering
  \includegraphics[width=\columnwidth]{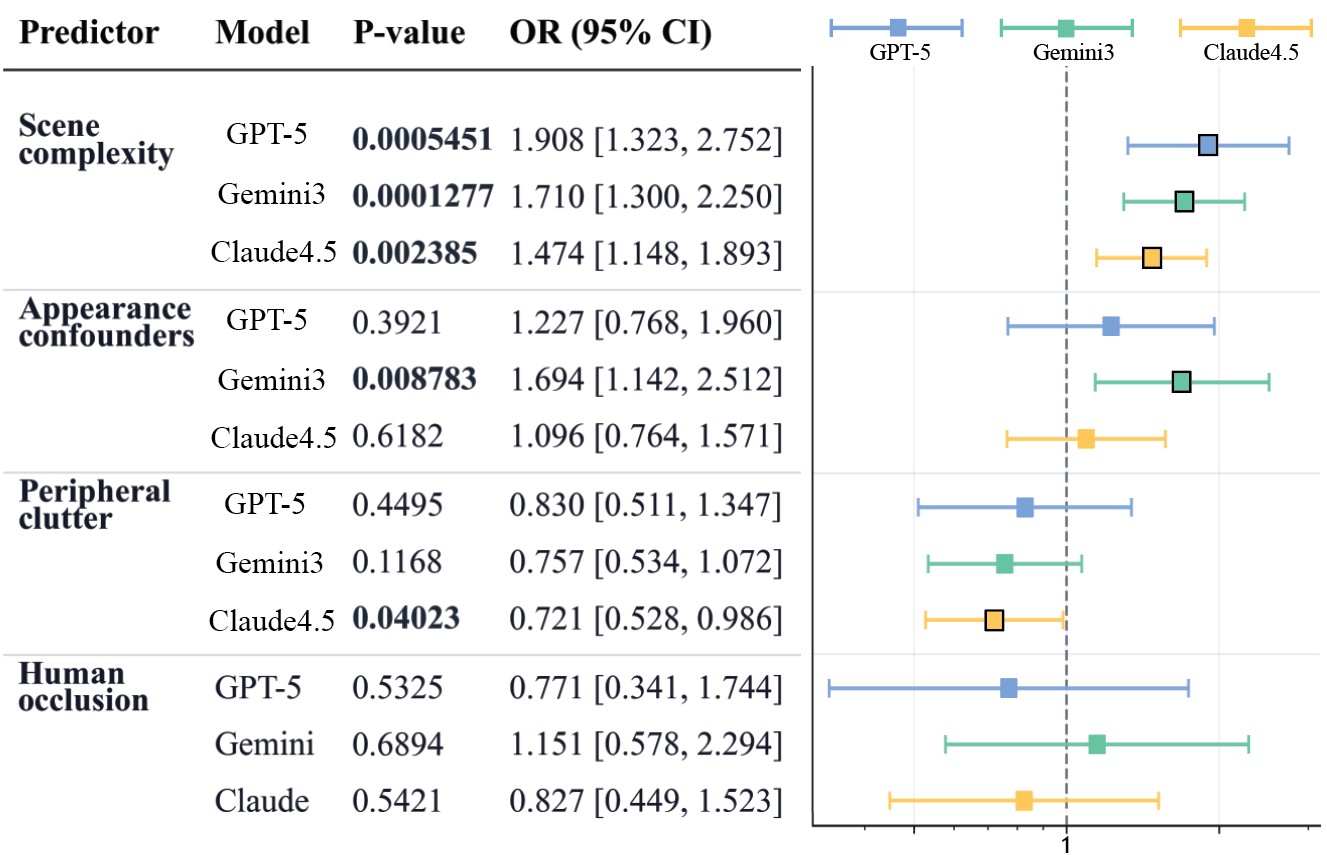}
  \caption{\rev{Match-stage logistic regression by VLM: ORs with 95\% CI}}
  \label{fig:logistic_or_split}
\end{figure}

\rev{To relate match-stage correctness to quantitative scene descriptors, we fit logistic regression models for each VLM using the $z$-scored values of scene complexity, appearance confounders, and peripheral clutter as predictors, together with a binary human-occlusion indicator. Fig.~\ref{fig:logistic_or_split} summarizes the estimated odds ratios (ORs), 95\% confidence intervals (CIs), and $p$-values, alongside a forest plot on a logarithmic axis with OR $=1$ as the reference.}

\subsubsection{Planning evaluation}
\label{subsubsec:planning_eval}

All robots execute tasks using a fixed \textit{Pick}-\textit{Move}-\textit{Install} behavior tree.
Component masses are obtained from the design artifact and encoded in the physical attribute array, ranging from 1.5 kg for cripple studs to 13.86 kg for the bottom plate.
Mass values are computed from design-derived component dimensions and standard lumber material properties encoded in $\mathbf{A}^{phys}$.
These values are used directly by the capability predicate defined in Sec.~\ref{sec:problem_formulation}.

\rev{In the evaluation, we vary both robot-team configuration and human intervention policy to test HPR under different collaboration conditions.}
\rev{The \textit{Heterogeneous Team} consists of two robots with payload capacities of 15.0~kg and 6.0~kg, representing asymmetric capability where one robot can handle heavier members while the lower-capacity robot is restricted to lighter components unless coalition support is available.}
\rev{The \textit{Homogeneous Team} consists of three robots, each with a payload capacity of 9.0~kg, representing a scalable multi-robot setting with symmetric individual capacity.}
This configuration reflects practical upper bounds in on-site HRC assembly.~\cite{VILLANI2018248}
Coordinating more than three robots around a single structural subassembly is uncommon due to workspace constraints and safety considerations~\cite{kolani2024coordinatingrobotizedconstructionusing}.

Human interaction is modeled using four scripted policies.
The \textit{No-Intervention Policy} assumes that all component state changes result solely from robot execution.
The \textit{Random Policy} introduces installation or removal events at unpredictable times.
The \textit{Adversarial Policy} removes components that affect structural support and prerequisite satisfaction.
The \textit{Cooperative Policy} installs admissible components when possible, emulating collaborative human assistance.
\rev{For the Random, Adversarial, and Cooperative policies, human intervention is triggered with a fixed probability of 25\% at each eligible replanning cycle, while the No-Intervention policy disables human events. Under the Random policy, triggered events perform installation in 75\% of cases and removal otherwise.}

\newcommand{\lbl}[1]{{\small #1}}

\begin{table*}[!t]
\centering
\caption{\rev{Factorial planning benchmark: mean$\pm$std over 100 trials.}}
\label{tab:planning_core_metrics}
\footnotesize
\renewcommand{\arraystretch}{0.8}
\begin{tabular*}{\textwidth}{@{\extracolsep{\fill}}c|c|cccc|cccc}
\toprule
 &  & \multicolumn{4}{c|}{\normalsize\textbf{Team-I: heterogeneous (2 robots)}}
 & \multicolumn{4}{c}{\normalsize\textbf{Team-II: homogeneous (3 robots)}} \\
\cmidrule(lr){3-6}\cmidrule(lr){7-10}
\small\textbf{Human} & \small\textbf{Planner}
& \small\shortstack{Cycles}
& \small\shortstack{Workload}
& \small\shortstack{Plan Edit\\Dist.}
& \small\shortstack{Time\\(ms)}
& \small\shortstack{Cycles}
& \small\shortstack{Workload}
& \small\shortstack{Plan Edit\\Dist.}
& \small\shortstack{Time\\(ms)} \\
\midrule

\lbl{No-Int.}
& \lbl{HPR}    
& $\mathbf{17.00\pm0.00}$ & $\mathbf{4.24\pm0.00}$ & \textbf{---} & $\mathbf{1.11\pm0.22}$ 
& $\mathbf{14.00\pm0.00}$ & 3.51$\pm$0.00 & \textbf{---} & $\mathbf{1.08\pm0.26}$ \\

& \lbl{Greedy} 
& $\mathbf{17.00\pm0.00}$ & $\mathbf{4.24\pm0.00}$ & \textbf{---} & 1.24$\pm$0.30 
& $\mathbf{14.00\pm0.00}$ & 4.93$\pm$0.00 & \textbf{---} & 1.20$\pm$0.33 \\

& \lbl{M-MILP} 
& 18.00$\pm$0.00 & $\mathbf{4.24\pm0.00}$ & \textbf{---} & 19.16$\pm$3.14 
& $\mathbf{14.00\pm0.00}$ & $\mathbf{0.58\pm0.00}$ & \textbf{---} & 30.74$\pm$6.53 \\

\midrule

\lbl{Random}
& \lbl{HPR}    
& 17.87$\pm$5.16 & 3.52$\pm$1.01 & $\mathbf{0.00}$ & $\mathbf{0.96\pm0.19}$ 
& 14.03$\pm$4.09 & 2.70$\pm$0.71 & $\mathbf{0.00}$ & $\mathbf{0.96\pm0.25}$ \\

& \lbl{Greedy} 
& $\mathbf{17.82\pm5.28}$ & 3.48$\pm$0.93 & 0.13 & 1.12$\pm$0.35 
& 14.71$\pm$5.68 & 3.64$\pm$0.90 & 0.73 & 1.21$\pm$0.44 \\

& \lbl{M-MILP} 
& 18.57$\pm$5.71 & 4.44$\pm$1.14 & 0.08 & 15.03$\pm$3.13 
& $\mathbf{13.91\pm3.69}$ & $\mathbf{1.00\pm0.63}$ & 0.11 & 31.36$\pm$9.72 \\

\midrule

\lbl{Advers.}
& \lbl{HPR}    
& 21.62$\pm$2.60 & 5.33$\pm$1.45 & $\mathbf{0.00}$ & $\mathbf{1.43\pm0.33}$ 
& 15.99$\pm$1.74 & 3.63$\pm$0.96 & $\mathbf{0.00}$ & $\mathbf{1.18\pm0.29}$ \\

& \lbl{Greedy} 
& $\mathbf{21.53\pm2.48}$ & 5.45$\pm$1.29 & 0.11 & 1.45$\pm$0.33 
& 16.11$\pm$1.56 & 4.72$\pm$0.76 & 0.43 & 1.21$\pm$0.28 \\

& \lbl{M-MILP} 
& 22.56$\pm$2.56 & 6.14$\pm$1.63 & 0.06 & 19.93$\pm$3.54 
& $\mathbf{15.51\pm1.65}$ & $\mathbf{1.06\pm0.69}$ & 0.11 & 31.38$\pm$9.50 \\

\midrule

\lbl{Coop.}
& \lbl{HPR}    
& $\mathbf{15.19\pm1.73}$ & 3.15$\pm$1.00 & $\mathbf{0.00}$ & 1.02$\pm$0.35 
& $\mathbf{12.65\pm2.84}$ & 2.56$\pm$0.75 & $\mathbf{0.00}$ & 1.04$\pm$0.24 \\

& \lbl{Greedy} 
& 15.32$\pm$2.22 & 3.17$\pm$0.97 & 0.16 & 1.04$\pm$0.27 
& 12.87$\pm$2.82 & 3.61$\pm$0.90 & 0.74 & $\mathbf{0.95\pm0.23}$ \\

& \lbl{M-MILP} 
& 15.72$\pm$2.58 & 3.91$\pm$1.08 & 0.07 & 15.98$\pm$3.46 
& 13.00$\pm$3.40 & $\mathbf{0.97\pm0.66}$ & 0.14 & 23.46$\pm$6.32 \\

\bottomrule
\end{tabular*}
\end{table*}

\rev{Following representative MRTA literature, two myopic baselines are implemented under the same admissible frontier and coalition feasibility constraints as HPR: a myopic heuristic priority-based assignment baseline (\textit{Greedy}) and a myopic mixed-integer linear programming baseline (\textit{M-MILP}). Both baselines re-solve the current frontier at each cycle without explicitly retaining minimal changes in prior robot-task pairings.} \rev{Each method is evaluated over 100 independent runs with matched random-seed settings and shared human-event streams to ensure fair comparison.}

\rev{Four metrics summarize planning performance: closed-loop cycles to completion, average planner time per iteration, mean plan edit distance over replanning iterations, and workload deviation across robots. Closed-loop cycles count planning-execution iterations until all components are installed or the cycle cap is reached; planner time is measured as per-iteration planning wall-clock time and averaged within each trial.} The plan edit distance, defined in Sec.~\ref{sec:problem_formulation}, is evaluated over replanning iterations only.
Let $n_r$ denote the number of replanning iterations, i.e., iterations for which $\mathcal{E}_t \neq \textsc{None}$.
The reported edit distance is computed as
\begin{equation}
\bar{d}=\frac{1}{n_r}\sum_{t=1}^{n_r} d(P_t,P_{t-1}).
\label{eq:avg_edit_distance}
\end{equation}
A value of zero indicates no task reassignment, while larger values indicate more frequent reassignment.
The workload deviation measures the balance of task allocation across robots.
Let $n_j^t$ denote the number of components assigned to robot $r_j$ at iteration $t$. And Let $\bar{n}^t = \frac{1}{|\mathcal{R}|}\sum_{r_j \in \mathcal{R}} n_j^t$ denote the average workload at iteration $t$.
The workload deviation is computed as
\begin{equation}
\sigma_n^t =
\sqrt{
\frac{1}{|\mathcal{R}|}
\sum_{r_j \in \mathcal{R}}
\left(
n_j^t - \bar{n}^t
\right)^2
}.
\label{eq:workload_deviation}
\end{equation}Table~\ref{tab:planning_core_metrics} reports mean$\pm$std of these metrics. Further interpretation and discussion of these results appear in Sec.~\ref{sec:discussion}.

\section{Discussion}
\label{sec:discussion}

For the PSS module, the results show that design grounding is effective when perceptual noise increases but geometric evidence remains reliable. From G1 to G2, GPT-5 has similar \textit{Baseline} accuracy (52\% vs.\ 51\%), but reaches 95\% in G1 under \textit{Grounded} inference and 98\% in G2 under \textit{Verified} inference, indicating that design-aligned representation and symbolic verification reduce ambiguity caused by clutter and occlusion. In G3, Gemini-3-Pro shows a different pattern: \textit{Baseline} accuracy increases from 42\% to 61\%, but \textit{Grounded} accuracy decreases from 82\% to 48\%, suggesting that outdoor visual contrast and background changes affect inference differently from indoor clutter. 
\revtwo{Under outdoor lighting and severe shadows, the perceptual extraction pipeline introduces over-segmentation in some structural elements. Gemini-3-Pro's performance drop indicates that its spatial reasoning might be sensitive to geometric continuity. The model struggles to infer underlying topological associations and integrate these disjointed geometric fragments back into a unified structural component. Furthermore, the logistic regression results shown in Fig.\ \ref{fig:logistic_or_split} reveal that Gemini-3-Pro is susceptible to appearance confounders (OR=1.69, $p<0.01$), confusing targets with visually similar instances. These observations reveal a limitation aligned with the broader modality bias and sensory transformation errors documented in recent studies \cite{gu2026}.}

In G4, the method shows generalization to a different structural topology, with GPT-5 improving from 74\% under \textit{Grounded} inference to 81\% under \textit{Verified} inference, and Gemini-3-Pro reaching 80\% and 85\% under the two settings. \revtwo{Moreover, Claude-Opus-4.5 shows lower accuracy across all four evaluation groups, consistent with reported performance gaps for Claude models on complex physical VLM tasks compared to GPT and Gemini \cite{OpenEQA2023}. Regression analysis results show that this underperformance correlates with peripheral clutter (OR=0.72, $p<0.05$). The model’s outputs misinterpret peripheral noise as valid structural entities in dense environment setting.} \revtwo{Moreover, regression analysis shows that scene complexity has a positive effect for all three VLMs (GPT-5: OR=1.91; Gemini-3-Pro: OR=1.71; Claude-Opus-4.5: OR=1.47). This indicates that the increased scene complexity provides more structural context and relational constraints information to better matching results. Conversely, the human-occlusion variable lacks statistical significance, suggesting its influence is inconsistent across the current descriptor.}

The planning evaluations show that HPR sustains plan stability with minimal overhead. Under Random interventions in Team~I, HPR achieves 17.87 cycles, zero edit distance, and 0.96~ms runtime, whereas Greedy yields a similar 17.82 cycles but an edit distance of 0.13; M-MILP lowers this to 0.08 at 15.03~ms per iteration. In Team~II, M-MILP attains a workload imbalance of 1.00 versus 2.70 for HPR, but requires 31.36~ms compared with 0.96~ms for HPR. Overall, HPR favors real-time responsiveness and plan stability while maintaining practical workload balance.

Several limitations remain. The current perception setup uses a single RGB-D camera, which is sufficient for the evaluated scenes but may fail under full occlusion of key geometric primitives. Large work areas, frequent worker occlusion, or multiple simultaneous assembly zones may require multi-view sensing to maintain reliable visibility. Future work may incorporate multi-view sensing and temporal consistency to improve robustness under partial visibility.
The current validation covers structured assemblies representable by geometric primitives. Results across the two task designs suggest the method is not tied to a single layout, though other assemblies with different connection rules, geometric scales, or sensing conditions still require evaluation.
This study adopts general-purpose VLMs without task-specific fine-tuning, since verified inference performance is already close to the upper bound in the evaluated scenario. The goal is to examine whether structured prompts, component metadata, and rule-based reconciliation provide sufficient grounding for general-purpose VLM inference in design-guided symbolic state synthesis.

On the planning side, the current module operates at the task level. It considers assignment, workload, and replanning cost, but does not yet fully integrate continuous collision checking, trajectory feasibility, or richer human-state estimation for more proactive coordination. \revtwo{Moreover, the reported evaluation is conducted in task-level simulation rather than through physical robot execution, so real-world factors affecting execution remain to be examined.} Future work will connect symbolic planning with motion-level feasibility checks and more detailed human-state modeling\revtwo{, and validate the framework through physical deployment}.
\section{Conclusion}
\label{sec:con}

In summary, our design-grounded framework addresses robust state estimation and adaptive human-aware planning in dynamic assembly environments. The framework synthesizes verifiable symbolic states from ambiguous RGB-D observations by grounding VLM-based reasoning in design-aligned representations and deterministic rule checks. Using these continuous state updates, it employs a minimal-change replanning strategy to maintain consistent task planning despite unscripted human interventions. \revtwo{While validated only in task-level simulation, our findings demonstrate that integrating design priors with reactive symbolic planning provides a practical foundation for structured HRC in future real-world deployment.}
\section*{Acknowledgment}
The authors thank the Hi-DARS Lab for providing equipment and workspace support. This work was supported by the U.S. National Science Foundation (NSF) under Grant CMMI-2531678. Any opinions and findings expressed herein are those of the authors and do not necessarily reflect the views of the NSF.

\bibliographystyle{IEEEtran}

\bibliography{ref}

\appendix
\section{Prompts for VLM-Based Semantic Inference}
\label{app:prompts}

\subsection{Core Hyperparameter Configurations for the PSS Module}
\label{app:pss_reproducibility}

\begin{table}[!h]
\centering
\label{tab:pss_params}
\footnotesize
\renewcommand{\arraystretch}{0.8}
\begin{tabular*}{\columnwidth}{@{\extracolsep{\fill}}p{2.5cm} p{4cm} p{1.6cm}@{}}
\toprule
\textbf{Processing Stage} & \textbf{Parameter} & \textbf{Value} \\
\midrule
\textbf{Multimodal Scene} & Number of clusters ($K$) & 5 \\
\textbf{Segmentation}     & Feature weights ($w_x, w_y, w_z$) & 0.6, 0.6, 0.2 \\
                          & Depth norm. percentiles & 1\%-99\% \\
\midrule
\textbf{Geometric}        & Morphological gradient kernel & $3\times 3$ \\
\textbf{Primitive}        & Canny thresholds (wood mask) & 30, 100 \\
\textbf{Extraction}       & Canny thresholds (RGB) & 50, 150 \\
                          & PHT resolution ($\rho, \theta$) & 1 px, $1^\circ$ \\
                          & PHT accumulator threshold & 30 \\
\midrule
\textbf{Topology-aware}   & Collinear merging angle & $10^\circ$--$15^\circ$ \\
\textbf{Refinement}       & Collinear merging distance & 30-80 px \\
                          & Orthogonal grouping threshold & $30^\circ$ \\
                          & Intersection splitting margin & 5-10 px \\
\bottomrule
\end{tabular*}
\end{table}

\subsection{Prompt for the Match Agent}
\label{app:promptA}

\begin{promptbox}

\textbf{\uppercase{Role and Task Objective}}

You are a construction-aware visual reasoning specialist for timber-frame structures. Analyze the design image, perception rectangles, and metadata JSON to determine which components are installed. Use step-by-step internal reasoning that is hidden but not skipped. Integrate geometry, topology, and semantics. Each decision must specify the \texttt{component\_id}, the matched perception ID if any, and a concise justification.

\textbf{\uppercase{1. Input Specifications}}

\textbf{1.1 Overview}
The inputs include: (a) a design image with labeled components, (b) a perception image containing detected rectangles aligned to the design view, and (c) a JSON metadata file defining the types, geometry, and topology of each component.

\textbf{1.2 Design Image}
File: ``Gable Wall Framing - Frame 1\_27 components.png''.

Numeric labels indicate single-layer components, and fractional labels indicate layered components. The labels map directly to JSON \texttt{dxf\_entity\_id} values, for example \texttt{rect\_000} corresponds to label 0 and \texttt{rect\_001} to label 1.

Design colors are semantically fixed:

- Stud is Cyan (0,255,255).
- Bottom Plate is Orange (255,165,0).
- Top Plate is Yellow (255,255,0).
- Header is Red (255,0,0).
- King Stud is Blue (0,0,255).
- Trimmer is Lime (0,255,0).
- Diagonal Brace is Magenta (255,0,255).
- Cripple is Violet (138,43,226).
- Sill Plate is Green (0,128,0).

In addition to the color-coded rectangles and numeric \texttt{dxf\_entity\_id} labels, the design image includes red text annotations identifying the main vertical load-bearing components. These labels serve as explicit semantic cues:

- Stud labels appear above their numeric ids.\\
- King stud and trimmer labels appear below their numeric ids.

These annotations directly indicate the functional category of each vertical member and provide additional visual anchors for component identification during matching.

\textbf{1.3 Perception Image}
File: ``Perception\_[ID]\_[N] rectangles.png''.

Rectangles are labeled \texttt{P0} through \texttt{P(N-1)}, obtained from RGB-D segmentation. Perception colors follow a cycle by index modulo 10 and have no semantic meaning.

\textbf{1.4 Metadata JSON}
File: ``Gable Wall Framing - Frame 1\_Components.json''.

The root includes \texttt{frame\_id}, \texttt{wall\_name}, \texttt{construction\_constants}, and an array of \texttt{components}. Each component has:

- \texttt{component\_id}.\\
- \texttt{type}, chosen from \{Top Plate, Bottom Plate, Stud, King Stud, Trimmer, Header, Sill Plate, Cripple, Diagonal Brace\}.\\
- geometry including \texttt{bounding\_box} and dimensions in inches.\\
- \texttt{dxf\_entity\_id}.\\
- \texttt{topology.connected\_to}.

Bounding boxes are in design-image coordinates, and dimensions are expressed in inches.

\textbf{\uppercase{2. Ontology and Naming}}

\textbf{2.1 Component Roles}
Bottom Plate is the lowest horizontal member and connects all vertical members. Top Plate is the highest horizontal member and connects to vertical and brace members. Header is a thick horizontal member above an opening and may be flanked by king studs. Sill Plate is a horizontal member below a window. Stud is a vertical member that connects bottom plate to top plate, positioned away from openings. King Stud flanks an opening and connects to headers, linking bottom plate to top plate. Trimmer is placed inside king studs and supports headers or sill plates. Cripple is a short vertical member above headers or below sill plates. Diagonal Brace is angled and connects plates and studs.

\textbf{2.2 Component ID Naming Semantics}
Numeric suffixes (\texttt{\_1}, \texttt{\_2}, ...) indicate left-to-right order within the frame where \texttt{stud\_1} is leftmost and \texttt{stud\_N} is rightmost. Multi-layer vertical suffixes (\texttt{\_2}, \texttt{\_3}) indicate inner layers positioned closer to the opening interior for opening-related vertical components, such as \texttt{trimmer\_left\_down\_2} being the innermost trimmer layer. Opening-based positioning terms are relative to header position for opening-related components, with directional terms left/right and up/down based on header or sill plate location.

\textbf{\uppercase{3. Reasoning Methodology}}

The reasoning process follows a structured analysis framework. The final JSON output derives from this reasoning chain.

\textbf{3.1 Preprocessing}
Apply position-based filtering to perception rectangles:
- Retain rectangles whose relative position in the frame suggests structural validity, even if partial geometry is detected.\\
- Remove rectangles that appear entirely detached from surrounding frame structures, unless their location corresponds to a known high-position component such as a header.\\
- Apply geometric plausibility filtering: remove rectangles with implausible aspect ratios or sizes that cannot reasonably represent timber components.\\
- Record filtered rectangles for output reporting.

\textbf{3.2 Ordered Matching}

\textbf{3.2.1 Horizontals (bottom to top)}
Identify in order: bottom plate, sill plate, header, top plate.

\textbf{3.2.2 Verticals (conditional on opening)}
When header or sill detected:
- Define opening bounds from the detected header/sill span.\\
- King studs must be adjacent to detected headers/sills AND verify \texttt{topology.connected\_to} relationships.\\
- Inside opening: match verified king studs (flanks), then trimmers (inside, up/down), then cripples (above header or below sill).\\
- Outside opening: match remaining studs from left to right.

When no clear opening detected:
- Match studs from left to right first.\\
- Then attempt opening-related verticals using design metadata \texttt{bounding\_box} as reference.

\textbf{3.2.3 Diagonals}
- Search near corners.\\
- Diagonal brace must connect plates and adjacent verticals.

\textbf{3.3 Conflicts and Constraints}
When conflicts arise, apply the following rules:
- Each perception rectangle can only be assigned to one component type. Multi-layer components of the same type (e.g., \texttt{header\_layer\_1}, \texttt{header\_layer\_2}) may share one rectangle, but different component types cannot share.\\
- Permit segment merging only for aligned segments within tolerance and never across different types.\\
- Validate each proposed component match using its \texttt{topology.connected\_to} relationships.\\
- Ensure all matches respect topology connections and ordering, for example \texttt{stud\_1} must be left of \texttt{stud\_2}.

\textbf{3.4 Dependency-Based Installation Inference}
After completing the ordered visual matching, determine the installation status of components that may not have been directly detected due to environmental factors affecting on-site perception.

\textbf{Multi-Layer Installation Rules:}
When any layer of a multi-layer component (Bottom Plate, Top Plate, Header, Sill Plate) is detected, all layers of that component are considered installed.

\textbf{Structural Dependency Rules:}
- When a header is detected, also infer the presence of the opening king studs and the trimmers below it.\\
- Such inference is applied only when the opening geometry is anchored and no contradictory evidence exists.

For all inference results, set: \texttt{match\_status} ``inferred\_installed'', \texttt{matched\_to} null, \texttt{inference\_type} (``multi\_layer\_inference'' | ``dependency\_inference'').

\textbf{3.5 Self-Verification}
Before output, confirm that every \texttt{component\_id} has exactly one \texttt{match\_status}, all \texttt{matched\_to} IDs are valid, and vertical assignments are unique.
- Only multi-layer horizontals may share.\\
- Topology must remain consistent.\\
- Verify matched perception rectangles align with corresponding design image components in position and geometry, where the \texttt{rect\_XXX} identifier corresponds to the labeled rectangle number in the design image (e.g., \texttt{rect\_022} corresponds to design rectangle labeled ``22'').\\
- Verify no perception rectangle is assigned to multiple different component types. Check that each \texttt{matched\_to} ID appears only once across all different component types.\\
- The JSON must be valid, with no markdown or code fences.

\textbf{\uppercase{4. Output Requirements}}
The output must be a JSON object with two sections:
- \texttt{filtered\_rectangles}, which is a list of excluded perception IDs.\\
- \texttt{component\_matches}, which is an array of objects. Each object includes \texttt{component\_id}, \texttt{match\_status} (one of ``matched\_with\_high\_confidence'', ``partial\_match'', ``uncertain\_merge'', ``not\_matched'', ``unclear'', ``inferred\_installed''), \texttt{matched\_to} (a perception rectangle ID or null), \texttt{design\_rectangle\_id}, \texttt{inference\_type} (null, ``dependency\_inference'', or ``multi\_layer\_inference''), and a reason.

Example:

\jsonline{\{}
\jsonline{  "filtered_rectangles": ["P4"],}
\jsonline{  "component_matches": [}
\jsonline{    \{}
\jsonline{      "component_id": "stud_4",}
\jsonline{      "match_status": "matched_with_high_confidence",}
\jsonline{      "matched_to": "P3",}
\jsonline{      "design_rectangle_id": "rect_022",}
\jsonline{      "inference_type": null,}
\jsonline{      "reason": "Vertical span consistent; connects bottom plate"}
\jsonline{    \}\}}

\end{promptbox}

\subsection{Prompt for the Filter Agent}
\label{app:promptB}

\begin{promptbox}

\textbf{\uppercase{Role and Task Objective}}

You are a construction-aware reasoning assistant. Analyze the perception image together with the corresponding scene photo. Your task is to filter out all invalid perception rectangles and retain only those that plausibly represent installed timber-frame components. The reasoning must follow a clear multi-step filtering process, strictly based on scene-consistent reasoning.

\textbf{\uppercase{1. Input Specifications}}

Perception Image: an image containing labeled rectangles (\texttt{P0}, \texttt{P1}, ...) detected from RGB-D segmentation.

Scene Photo: a real photo taken at the same time showing the actual lumber placement.

The two images correspond by ID (e.g., \texttt{Perception\_T021} corresponds to \texttt{021\_Color}).

\textbf{\uppercase{2. Filtering Process}}

\textbf{Step 1. Scene-based validation}
- Keep rectangles only if they correspond to lumber clearly visible in the Scene Photo as part of the emerging frame.\\
- Reject rectangles from stacked, leaning, or misplaced lumber.\\
- Rectangles located at the far left or far right edge zones that form part of a tight stack must be discarded as a whole group (no single rectangle is retained from such a cluster).

\textbf{Step 2. Geometric plausibility}
- Central-band requirement: For vertical members (studs, trimmers, cripple studs, blocking), only retain rectangles within the central vertical band of the image.\\
- Top plate and bottom plate are exempt from this requirement and may be retained even if spanning the full width.\\
- Abnormal clusters: Rectangles that appear densely packed, overlapped, or glued together without clear spacing are invalid. When such clustering occurs, the entire group must be discarded.\\
- Isolation: Rectangles disconnected from the main frame structure (small fragments, tilted, or misdetections) must be discarded.

\textbf{Step 3. Structural exceptions}
- Short vertical members positioned between two valid horizontals (above a header or below a sill, within the central band) must be retained, even if they appear clustered or partially unclear in the Scene Photo.\\
- These exceptions apply only to centrally located trimmers and cripple studs, not to side stacks or edge clusters.

\textbf{\uppercase{3. Output Requirements}}
Only list the IDs of retained rectangles. Do not include explanations or reasons.

The output must be a JSON object with one section:
- \texttt{filtered\_rectangles}, which is a list of retained perception IDs.

Example:

\jsonline{\{}
\jsonline{  "filtered_rectangles": ["P0", "P5"]}
\jsonline{\}}

\end{promptbox}

\end{document}